\documentclass[12pt]{spieman}
\usepackage{amsmath,amsfonts,amssymb}
\usepackage{bm}
\usepackage{verbatim}
\usepackage{booktabs}
\usepackage{algorithm} 
\usepackage{algorithmic} 
\usepackage{multirow,multicol} 
\usepackage{xcolor}
\usepackage{caption}
\usepackage{graphicx}
\usepackage{subfigure}
\usepackage{float}
\usepackage{multirow}
\usepackage{multicol} 
\usepackage{colortbl,booktabs}
\usepackage{setspace}
\usepackage[titles,subfigure]{tocloft}
\title{Cross-modal Subspace Learning via Kernel Correlation Maximization and Discriminative Structure Preserving}
\author[a,b]{Jun Yu}
\author[a,b,*]{Xiao-Jun Wu}
\affil[a]{The School of IoT Engineering, Jiangnan
University, 214122, Wuxi, China.}
\affil[b]{The Jiangsu Provincial Engineering Laboratory
of Pattern Recognition and Computational Intelligence, Jiangnan University
214122, Wuxi, China.}

\begin{document} 
\maketitle

\begin{abstract}The measure between heterogeneous data is still an open problem. Many research works have been developed to learn a common subspace where the similarity between different modalities can be calculated directly. However, most of existing works focus on learning a latent subspace but the semantically structural information is not well preserved. Thus, these approaches cannot get desired results. In this paper, we propose a novel framework, termed Cross-modal subspace learning via Kernel correlation maximization and Discriminative structure-preserving (CKD), to solve this problem in two aspects. Firstly, we construct a shared semantic graph to make each modality data preserve the neighbor relationship semantically. Secondly, we introduce the Hilbert-Schmidt Independence Criteria (HSIC) to ensure the consistency between feature-similarity and semantic-similarity of samples. Our model not only considers the inter-modality correlation by maximizing the kernel correlation but also preserves the semantically structural information within each modality. The extensive experiments are performed to evaluate the proposed framework on the three public datasets. The experimental results demonstrated that the proposed CKD is competitive compared with the classic subspace learning methods. 
\end{abstract}

\keywords{Cross-modal retrieval, subspace learning, kernel correlation, discriminative, HSIC}\\

%

\begin{spacing}{2}
\section{Introduction}
%
%
%
%
Recently, the fast development of the Internet and the explosive growth of multimedia including text, image, video, audio has greatly enriched people's life but magnified the challenge of information retrieval. Representative image retrieval methods, such as Region- based image retrieval \cite{MTAA:Zhang}, Color-based image retrieval \cite{JEI:Ciocca}, Contour Points Distribution Histogram(CPDH) \cite{IAVC:shuxin}, Inverse Document Frequency (IDF) \cite{IP:Zheng}, Content-based image retrieval \cite{MTAA:Memon}, are not directly applied in multimodal. Multimodal data refers to those data of different types but with the same content information, for example, recording video clips, music, photos and tweets of a concert. Cross-modal retrieval which aims to take one type of data as the query to return the relevant data of another type has attracted much attention.\\
\hspace*{0.35cm}The cross-modal retrieval methods need to solve a basic problem, i.e. how to measure the relevance between heterogeneous modalities. There are two strategies to solve this challenge. One is to directly calculate the similarity based on the known relationship between cross-modal data \cite{intro:Jia,intro:Jiang,intro:Song}. The other is to learn a latent common subspace where the distance between different modalities data can be measured, which is also termed as cross-modal subspace learning. The subspace learning models include unsupervised approaches \cite{{PIMPS:Akaho,CVPR2011:Sharma,NC:Tenenbaum}}, supervised approaches \cite{TPAMI:Kim,ACM:Rasiwasia,IEEEPAMI:Wang,CV:Gong,CVPR:Jacobs,ECCV:Lin}. Unsupervised methods usually use the intrinsical characteristic of data and the correlation between multimodal data to learn common subspace representation. There exist some inherent correlation among multiple modalities, since pair-wise multiple modalities describe the same semantic object. Canonical Correlation Analysis (CCA), kullback Leibler (KL) divergence, Hilbert-Schmidt Independence Criterion (HSIC) are widely used to measure the correlation between multiple modalities. However, unsupervised methods encounter bottlenecks and do not obtain satisfactory results bacause the available discriminative information is inadequate. To solve the problem, supervised methods introduce the semantic label information to learn the discriminative feature representation of each modality. Although existing supervised methods have achieved reasonable performance, they are still  some drawbacks. Most of these methods do not explore the semantically structural information and the correlation among multiple modalities simultaneously in the process of learning a latent subspace. Actually, the semantically structural information is very important to make the subspace representation more discriminative.\\
\hspace*{0.35cm}In this paper, we propose a novel framework which preserves both the shared semantic structure and the correlation among multiple modalites. The supervised information, semantic structure and multimodal data are collaboratively incorporated in a unified framework to uncover a common subspace for cross-modal retrieval, as illustrated in Fig. 1. The main contributions of our work are summarized as follows:\\
\hspace*{0.35cm}(1) The proposed learning model combines subspace learning with feature selection and semantic structure-preserving into a joint framework. A shared semantic graph is constructed to learn the discriminative feature representation of each modality. In addition, the convergence of the algorithm is analyzed theoretically.\\
\hspace*{0.35cm}(2) Extensive experiments are performed on three widely-used datasets and the experimental results show the superiority and effectiveness of our model.\\
\hspace*{0.35cm}Structurally, the rest of this paper falls into four parts. In Section 2, we review related works to cross-modal retrieval. In Section 3, we present the proposed CKD model and joint optimization process. The experimental results and analysis are provided in Section 4. In Section 5, we draw the conclusions of the paper.

\begin{figure}[htbp]
\centering
\includegraphics[width=\textwidth]{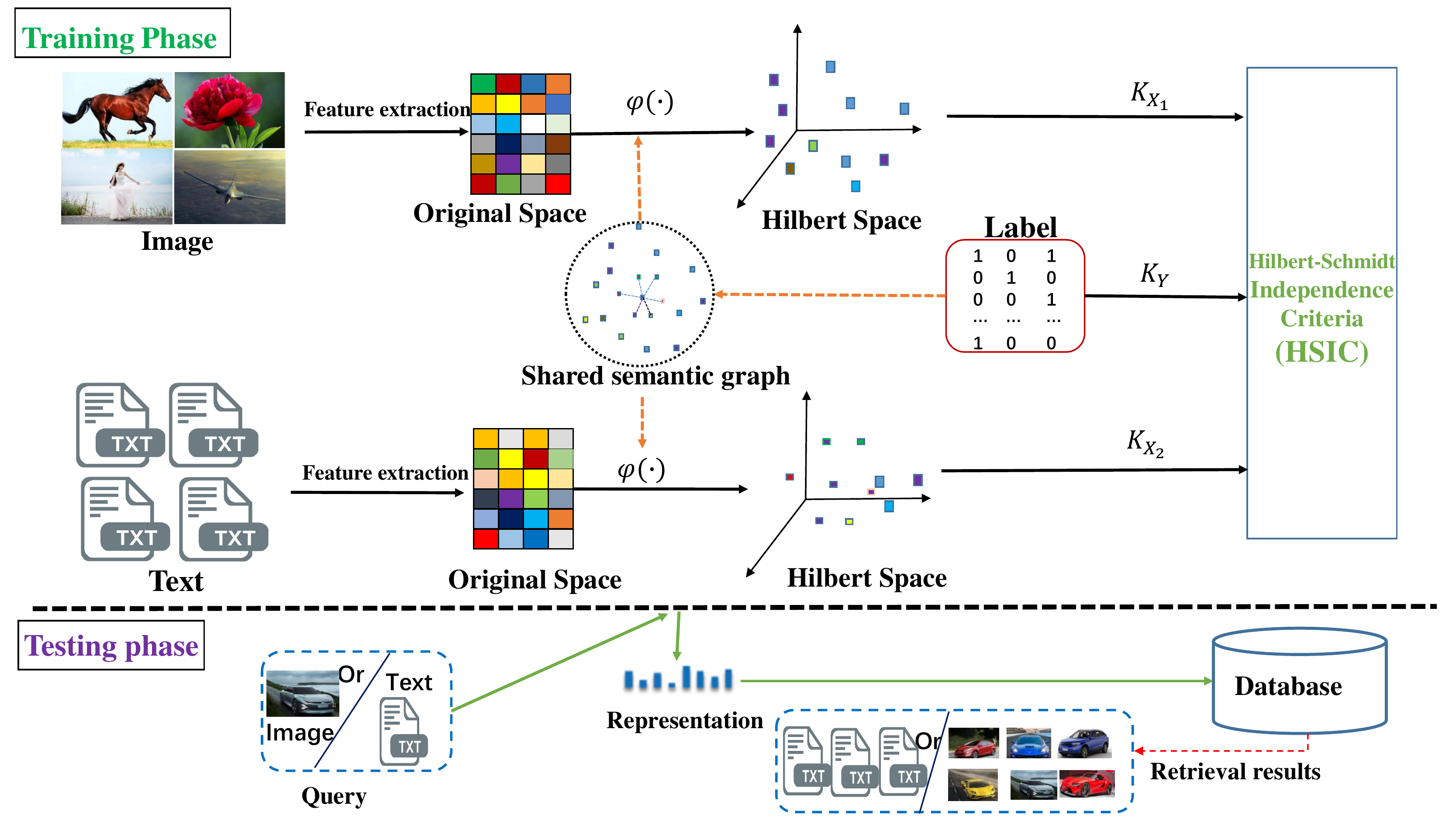}
\caption{The illustration of the model proposed in this paper. In the training phase, image modality and text modality are mapped to the common Hilbert space respectively. In Hilbert space, each modality preserves the semantically structural information. The HSIC is adopted to ensure the consistency of the similarity among data for each modality. In the testing phase, we can obtain the feature representation of an arbitrary query (an image or a text) in the learned Hilbert space, and other modal data that is similar content to the query is returned from the database.}
\label{Fig.1}
\end{figure}

\section{Related work}
Cross-modal retrieval has been a popular topic in multimedia applications. Many methods are developed to realize the retrieval tasks in the past decades. In this section, we preliminarily review the related works of this field. According to the different types of targeted representation, cross-modal retrieval methods are divided into two branches: real-valued and hash code representation learning. The hash code representation learning is referred to as cross-modal hashing. Cross-modal hashing approaches \cite{relatedworkhashing1,relatedworkhashing2,relatedworkhashing3,relatedworkhashing4,relatedworkhashing5,relatedworkhashing6} aim to learn a common Hamming space. Since cross-modal hashing methods are more geared towards retrieval efficiency, their focus is not on retrieval effectiveness. Real-valued representation learning approaches are designed to learn a common subspace where the similarity between different modalities can be measured directly.These approaches are roughly grouped into two paradigms: unsupervised subspace learning and supervised ones.

Unsupervised subspace learning methods explore the structure, distribution and topological information of data to learn common subspace representation for each modality. Canonical Correlation Analysis (CCA) \cite{relatedwork7} is the most representative unsupervised method. However, unsupervised methods do not obtain satisfying retrieval precision because the encoded features lack discriminative ability. Supervised methods exploit the semantic label information to learn a more discriminative common subspace. An example is three-view CCA (CCA-3V) \cite{relatedwork8} which regards the semantic label as a feature view to learn a common space for cross-modal retrieval. Multi-label CCA (ml-CCA) \cite{ICCV:Ranjan} is proposed to perform cross-modal retrieval by utilizing multi-label annotations. Semantic Correlation Matching (SCM) \cite{relatedwork9} uses the logistic regression to learn semantic subspace representation for each modality. Joint Feature Selection and Subspace Learning (JFSSL) \cite{relatedwork10} combines common subspace learning and coupled feature selection into a unified framework.

In recent years, many deep learning models have achieved promising performance. Deep CCA (DCCA) \cite{relatedwork11} extends the linear CCA to nonlinear one. Adversarial Cross-Modal Retrieval (ACMR) \cite{relatedwork12} is proposed to seek an effective common subspace based on adversarial learning. Wei et al proposed deep-SM \cite{relatedwork13} to learn a common semantic space where the probabilistic scores produced by a deep neural network are viewed as the semantic features. Deep Supervised Cross-Modal Retrieval (DSCMR) \cite{relatedwork14} aims to find a common representation space by minimizing the discrimination loss and modality invariance loss simultaneously. Adversarial Cross-Modal Embedding (ACME) \cite{relatedwork15} is proposed to resolve the cross-modal retrieval task in food domains.

The approach proposed in this paper employs the label information to learn effective and discriminative subspace representation. Since multimodal data appears in pairs and is semantically relevant, it is necessary to explore the correlation among multiple modalities. Besides, learned subspace representation should preserve the semantic structure, i.e. the closer the semantic relationship among samples the nearer the distance in common subspace and vice versa. Unlike most of the existing subspace learning methods, our model maximizes the correlation among multi-modal data and preserves the semantic structure within each modality simultaneously.
  
\section{Our Method}
\subsection{Problem Formulation}
Assume that $M$ modalities denotes as $X=\{X^{(1)},X^{(2)},...,X^{(M)}\}$. The $v$-th modality $X^{(v)}=\{X^{(v)}_1,X^{(v)}_2,...,X^{(v)}_n\}\in R^{n\times d_v}$($v=1,...,M$) contains $n$ samples with $d_v$ dimension. The label matrix is denoted by $Y=[y_1^T,y_2^T,...y_n^T]^T\in R^{n\times c}$, where $c$ is the number of categories.  $y_{ik}=1$ if the $i$-th sample belongs to the $k$-th class;  $y_{ik}=0$ otherwise. Without loss of generality, the samples of each modality are zero-centered, i.e. $\sum^n_{i=1}X^{(v)}_i=0$, ($v=1,2...M$). The aims of this paper is to learn isomorphic feature representation for heterogeneous multi-modal data.
\subsection{Hilbert-Schmidt independence criteria (HSIC)}
Given two mapping functions with $\phi(x):x\rightarrow R^d$ and $\phi(z):z\rightarrow R^d$. The associated positive definite kernel $K_x$ and $K_z$ are formulated as $K_x=<\phi(x),\phi(x)>$ and $K_z=<\phi(z),\phi(z)>$ respectively. The cross-covariance function between $x$ and $z$ is denoted as $C_{xz}=E_{xz}[(\phi(x)-u_x)\otimes(\phi(z)-u_z)]$, where $u_x$ and $u_z$ are the expectation of $\phi(x)$ and $\phi(z)$ respectively. The Hilbert-Schmidt norm of $C_{xz}$ is defined as $HSIC = \left \| C_{xz}\right \|_{HS}^2$. For $n$ paired data samples $D=\{(x_1,z_1),...,(x_n,z_n)\}$, An empirical expression of Hilbert-Schmidt independence criteria (HSIC) \cite{Neurocomputing:Xu,ICML:Davis,ITL:Principe} is defined as
\begin{equation}
  HSIC = (n-1)^{-2}tr(K_xHK_zH)
\end{equation}
where $H=\bm{I}-\frac{1}{n}\bm{1}_n\bm{1}_n^T$ is a centering matrix. The larger the $HSIC$, the stronger the correlation between $x$ and $z$.
\subsection{Model}
For simplicity, we discuss our algorithm based on two modalities, i.e. Image and Text. It is easy to extend to the case with more modalities.
\subsubsection{Kernel Correlation Maximization}
Our model adopts the HSIC to maximize the kernel dependence among multi-modal data. Multi-modal data are projected into the Hilbert space where we can calculate the kernel matrix $K_{X_v}=<V_v,V_v>=V_vV_v^T$, where $V_v=\phi_v(X^{(v)})=X^{(v)}P_v$ ($v=1,2$) and $P_v$ is the projection matrix of the $v$-th modal data. As the kernel matrix itself represents, in essence, the similarity relationship among samples, our model preserves the similarity relationship among samples for each modality, which is called the intra-modality correlation. Semantic label shared by multiple modalities is regarded as a semantic modality. Likewise, the kernel matrix of the semantic modality is signified as $K_Y=<Y,Y>=YY^T$. Our model preserves the consistence between feature-similarity and semantic-similarity of samples via maximizing kernel dependence. According to the definition (1), the objective formulation can be given as follows: 
\begin{equation}
\begin{split}
\max\limits_{P_1,P_2}&tr(HK_{X_1}HK_{X_2})+tr(HK_{X_1}HK_Y)+tr(HK_{X_2}HK_Y)\\& 
s.t.P_1^TP_1=I,P_2^TP_2=I
\end{split}
\end{equation}
where $K_{X_1}=X_1P_1P_1^TX_1^T, K_{X_2}=X_2P_2P_2^TX_2^T, K_Y=YY^T$. In this section, the inter-modality correlation and intra-modality similarity relationship are taken into account simultaneously in our model.
\subsubsection{Discriminative Structure-Preserving}
Although different modalities locate isomeric spaces, they share the same semantic information. We calculate the cosine similarity among samples by employing their semantic label vectors. Specifically, the similarity between the $i$-th and the $j$-th sample is defined as follows
\begin{equation}
S_{ij}=\frac{y_i\cdot y_j}{\rVert y_i\rVert_2\rVert y_j\rVert_2}
\end{equation}
where $\rVert y_i\rVert_2$ denote the $L_2$-norm of the vector $y_i$. We hope that the semantically structural relationship among samples is preserved in Hilbert space. That is to say, the closer the semantic relationship among samples the nearer the distance in the common Hilbert space. The problem can be formulated as the following objective function (4).
\begin{equation}
\begin{split}
\min\limits_{P_1,P_2}&\alpha_1tr(P_1^TX_1^TLX_1P_1)+\alpha_2tr(P_2^TX_2^TLX_2P_2)\\&
s.t. P_1^TP_1=I,P_2^TP_2=I
\end{split}
\end{equation}
where $L=diag(S\bm{1})-S$ denotes a graph Laplacian matrix and $\alpha_1$ and  $\alpha_2$ are two adjustable parameters. As discussed in some literature \cite{PR:Song} \cite{CSVT:Wang}, the $l_{2,1}$-norm constraint has some advantages of feature selection, sparsity, and robustness to noise. In our model, we impose the $l_{2,1}$-norm constraint on projection matrices to learn more discriminative features and remove the redundant features. The objective function (4) can be rewritten as follows
\begin{equation}
\begin{split}
\min\limits_{P_1,P_2}&\alpha_1(tr(P_1^TX_1^TLX_1P_1)+\lambda_1\rVert P_1\rVert_{2,1})\\&+\alpha_2(tr(P_2^TX_2^TLX_2P_2)+\lambda_2\rVert P_2\rVert_{2,1})\\&
s.t. P_1^TP_1=I,P_2^TP_2=I
\end{split}
\end{equation}
where $\lambda_1$ and $\lambda_2$ are two trade-off parameters.\\
 We integrate kernel dependence maximization and discriminative structure-preserving into a joint framework by combining (2) and (5). The overall objective function can be written as follows
\begin{equation}
\begin{split}
\min\limits_{P_1,P_2}&\alpha_1(tr(P_1^TX_1^TLX_1P_1)+\lambda_1\rVert P_1\rVert_{2,1})\\&+\alpha_2(tr(P_2^TX_2^TLX_2P_2)+\lambda_2\rVert P_2\rVert_{2,1})\\&+\beta(-tr(HK_{X_1}HK_{X_2})-tr(HK_{X_1}HK_Y)\\&\quad\quad-tr(HK_{X_2}HK_Y))\\&
s.t. P_1^TP_1=I,P_2^TP_2=I
\end{split}
\end{equation}
where $\beta$ is an adjustable parameter. We usually set $\beta=1$ for simplicity. The case with $\beta=0$ implies that our model only considers the discrimiantive structure-preserving.
\subsection{Optimization}
To optimize variables conveniently, we transform the $l_{2,1}$-norm constraint term into $tr(P_v^TD_vP_v)$ by adding an intermediate variable $D_v=diag(\frac{1}{\rVert P_v^{\cdot\cdot i}\rVert_2})$, where $P_v^{\cdot\cdot i}$ is the $i$-th row of $P_v(v=1,2)$. The objective function (6) can be rewritten as 
\begin{equation}
\begin{split}
\min\limits_{P_1,P_2}&-tr(HK_{X_1}HK_{X_2})-tr(HK_{X_1}HK_Y)\\&-tr(HK_{X_2}HK_Y)\\&
+\alpha_1(tr(P_1^TX_1^TLX_1P_1)+\lambda_1tr(P_1^TD_1P_1))\\&
+\alpha_2(tr(P_2^TX_2^TLX_2P_2)+\lambda_2tr(P_2^TD_2P_2))\\&
s.t. P_1^TP_1=I,P_2^TP_2=I
\end{split}
\end{equation}
where $K_{X_1}=X_1P_1P_1^TX_1^T, K_{X_2}=X_2P_2P_2^TX_2^T, K_Y=YY^T$.
\subsubsection{Optimization of $P_1$}Keeping only the terms relating to $P_1$, we can obtain
\begin{equation}
\begin{split}
\max\limits_{P_1}& tr(P_1^TQ_1P_1)\\&
s.t.P_1^TP_1=I
\end{split}
\end{equation}
where $Q_1=X_1^THX_2P_2P_2^TX_2^THX_1+X_1^THYY^THX_1-\alpha_1X_1^TLX_1-\alpha_1\lambda_1D_1$.
The optimal $P_1$ in (8) can be obtained via the eigenvalue decomposition on $Q_1$. 
\subsubsection{Optimization of $P_2$}Keeping only the terms relating to $P_2$, we can obtain
\begin{equation}
\begin{split}
\max\limits_{P_2}& tr(P_2^TQ_2P_2)\\&
s.t.P_2^TP_2=I
\end{split}
\end{equation}
where $Q_2=X_2^THX_1P_1P_1^TX_1^THX_2+X_2^THYY^THX_2-\alpha_2X_2^TLX_2-\alpha_2\lambda_2D_2$. Being similar to $P_1$, the optimal solution of $P_2$ can be got by the eigenvalue decomposition on $Q_2$.\\
\renewcommand{\algorithmicrequire}{\textbf{Input:}} 
\renewcommand{\algorithmicensure}{\textbf{Output:}}
\begin{algorithm}
\caption{The algorithm proposed in this paper (CKD)}
\label{Algorithm 1}
\begin{algorithmic}[1]
\REQUIRE The training data $X^{(v)}\in R^{n\times d_v}$; The label matrix $Y\in R^{n\times c}$; the dimension of the common Hilbert space $d$; Parameter $\alpha_v$ and $\lambda_v$. $(v=1,2)$
\ENSURE $P_1,P_2$\\
Initialize $P_1,P_2$.\\
Calculating similarity matrix $S$ according to (3).
\REPEAT
\STATE  Compute $D_1$ by $D_1^{ii}=\frac{1}{2\|P_1^{\cdot\cdot i}\|_2}$.\\
   \STATE  Compute $D_2$ by $D_2^{ii}=\frac{1}{2\|P_2^{\cdot\cdot i}\|_2}$.\\
    \STATE Update $P_1$ using Eq.(8)
	\STATE Update $P_2$ using Eq.(9)
\UNTIL Convergence \\
Return  $P_1,P_2$.
\end{algorithmic}
\end{algorithm}

\begin{figure*}[htbp]
\subfigure[Pascal-Sentence]{
\includegraphics[width=.3\textwidth]{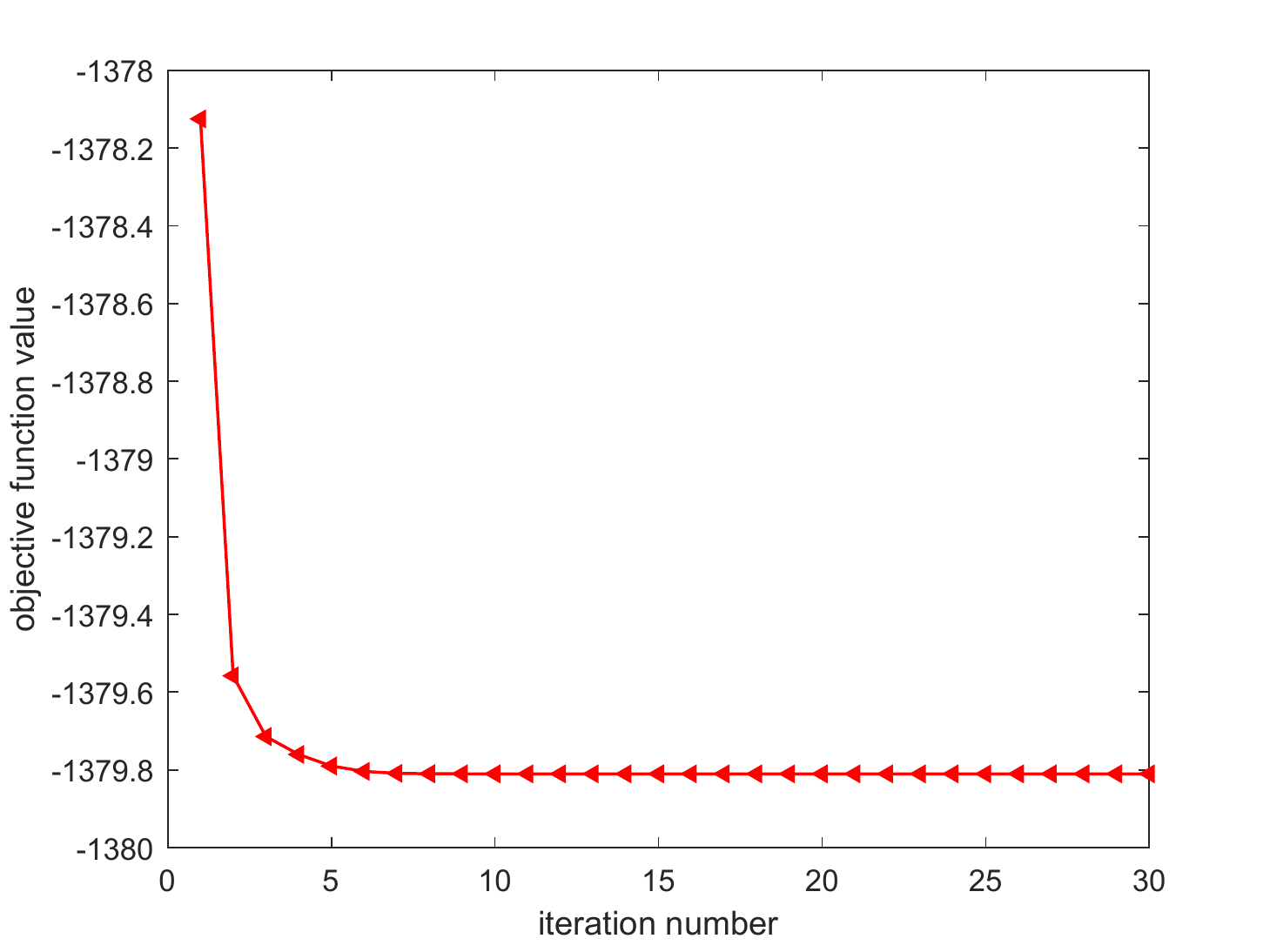}}
\subfigure[MIRFlickr]{
\includegraphics[width=.3\textwidth]{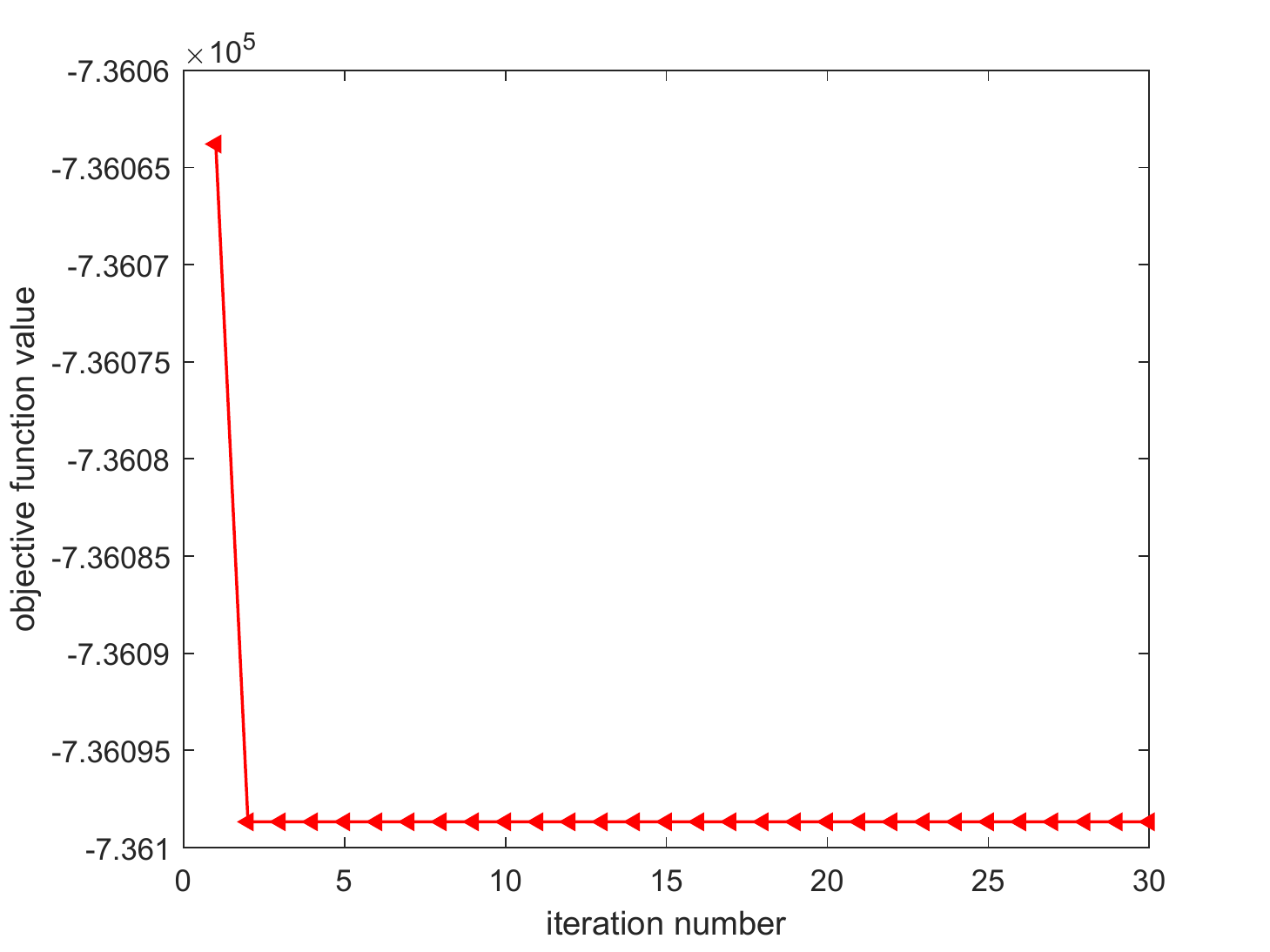}}
\subfigure[NUS-WIDE]{
\includegraphics[width=.3\textwidth]{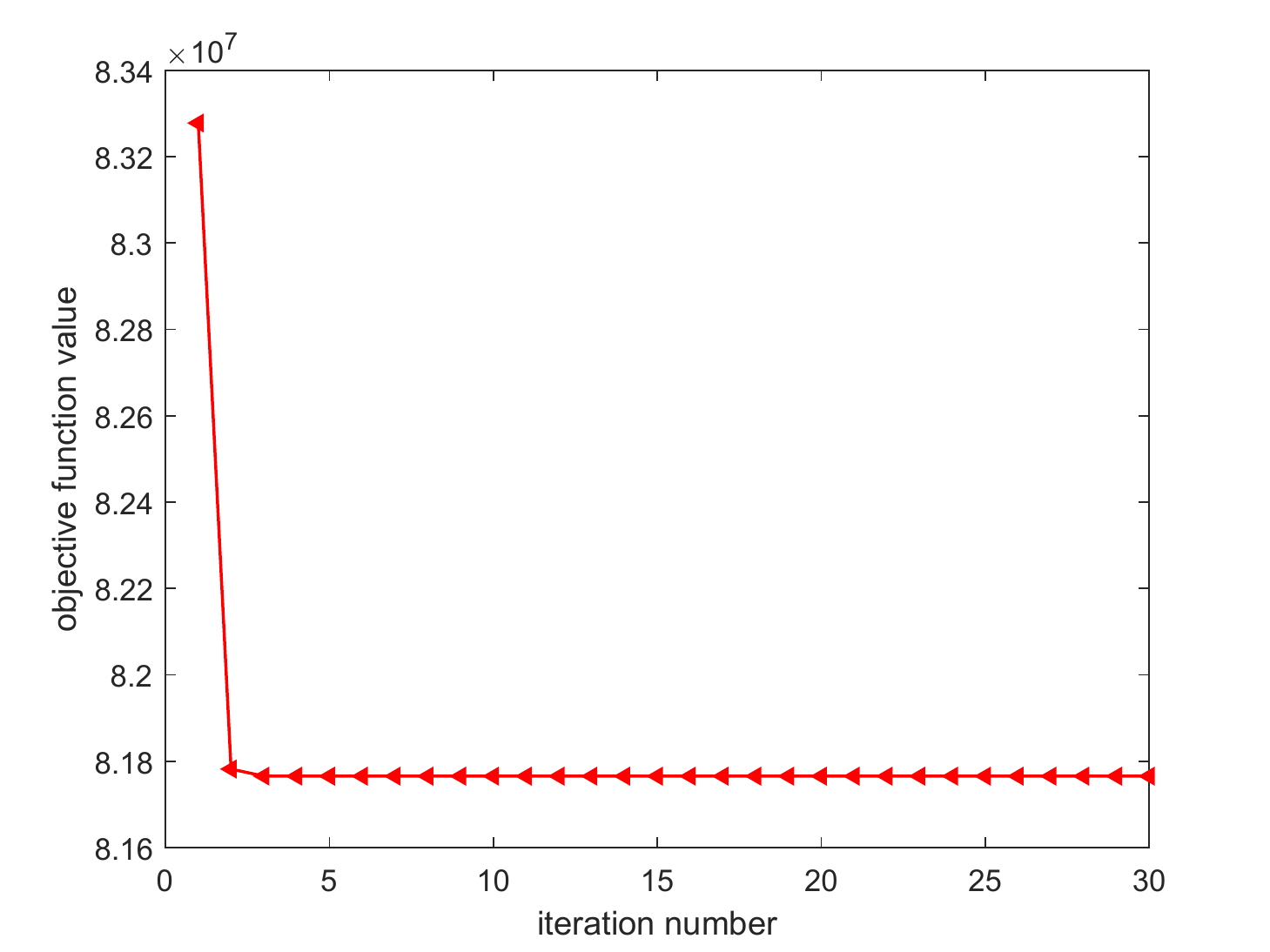}}
\caption{The convergence of algorithm 1 on Pascal-Sentence(a),MIRFlickr (b), and NUS-WIDE (c).}
\label{Fig.2}
\end{figure*}

\subsection{Convergence analysis}
The detail optimization procedure is summarized in Algorithm 1. The process is repeated until the algorithm converges. The convergence curves on NUS-WIDE, Pascal-Sentence and, MIRFlickr25k are plotted in Fig.2, which indicates that our method converges quickly.\\
$\bm{Theorem 1:}$ The objective function (7) based on the optimizing rule (Algorithm 1) is decreasing monotonically, and it converges to the global minimum value.\\
$\bm{Lemma 1:}$ Given any nonzero $f\in R^m$ and $g\in R^m$, then the following formula (10) holds. Please refer to \cite{ANIP:Nie} for details.
\begin{equation}
\frac{\rVert f\rVert_2}{2\rVert g\rVert_2}-\rVert f\rVert_2\ge\frac{\rVert g\rVert_2}{2\rVert g\rVert_2}-\rVert g\rVert_2
\end{equation}
In order to prove the $\bm{Theorem 1}$, we introduce the $\bm{Lemma 1}$. The detail proof about $\bm{Theorem 1}$ is given as follows.\\
 $\bm{Proof:}$ The optimization of $P_1$ and  $P_2$ are symmetrical in Algorithm 1, thus we just consider to prove one of them. The detailed proof with respect to the optimization of $P_1$ is provided below. The optimization problem about $P_1$ is written as follows
\begin{equation}
\begin{split}
\min\limits_{P_1}&-tr(HK_{X_1}HK_{X_2})-tr(HK_{X_1}HK_Y)\\&
+\alpha_1(tr(P_1^TX_1^TLX_1P_1)+\lambda_1tr(P_1^TD_1P_1))\\&
s.t. P_1^TP_1=I
\end{split}
\end{equation}

Letting $W(P_1)=tr(HK_{X_1}HK_{X_2})+tr(HK_{X_1}HK_Y)-\alpha_1tr(P_1^TX_1^TLX_1P_1)$, then the object function in Eq.(11) becomes $T(P_1)=\lambda_1tr(P_1^TD_1P_1)-W(P_1)$. For the $k$-th iteration,
\begin{equation}
\begin{split}
P_1^{(k+1)}&=\mathop{\arg\min}_{P_1}(\lambda_1tr(P_1^TD_1P_1)-W(P_1))\\&
\Rightarrow \lambda_1tr(P_1^{(k+1)^T}D_1^{(k)}P_1^{(k+1)})-W(P_1^{(k+1)})\\&
\leq \lambda_1tr(P_1^{(k)^T}D_1^{(k)}P_1^{(k)})-W(P_1^{(k)})\\&
\Rightarrow \lambda_1\sum_i^{d_1}\frac{\rVert P_1^{\cdot\cdot i(k+1)}\rVert_2^2}{2\rVert P_1^{\cdot\cdot i(k)}\rVert_2^2}-W(P_1^{(k+1)})\\&
\leq \lambda_1\sum_i^{d_1}\frac{\rVert P_1^{\cdot\cdot i(k)}\rVert_2^2}{2\rVert P_1^{\cdot\cdot i(k)}\rVert_2^2}-W(P_1^{(k)})\\&
\Rightarrow \lambda_1(\sum_i^{d_1}\frac{\rVert P_1^{\cdot\cdot i(k+1)}\rVert_2^2}{2\rVert P_1^{\cdot\cdot i(k)}\rVert_2^2}-\rVert P_1^{\cdot\cdot i(k+1)}\rVert_{2,1})\\&+\lambda_1\rVert P_1^{\cdot\cdot i(k+1)}\rVert_{2,1}-W(P_1^{(k+1)})\\&
\leq \lambda_1(\sum_i^{d_1}\frac{\rVert P_1^{\cdot\cdot i(k)}\rVert_2^2}{2\rVert P_1^{\cdot\cdot i(k)}\rVert_2^2}-\rVert P_1^{\cdot\cdot i(k)}\rVert_{2,1})\\&+\lambda_1\rVert P_1^{\cdot\cdot i(k)}\rVert_{2,1}-W(P_1^{(k)}).\\&
\end{split}
\end{equation}
By virtue of $\bm{Lemma 1}$, we can obtain
\begin{equation}
\lambda_1\rVert P_1^{\cdot\cdot i(k+1)}\rVert_{2,1}-W(P_1^{(k+1)})\leq \lambda_1\rVert P_1^{\cdot\cdot i(k)}\rVert_{2,1}-W(P_1^{(k)})
\end{equation}
then
\begin{equation}
\Rightarrow T(P_1^{(k+1)})\leq T(P_1^{(k)})
\end{equation}
Similarly, we can also prove that $T(P_2^{(k+1)})\leq T(P_2^{(k)})$ for the optimization of $P_2$. So, we have $T(P_1^{(k+1)},P_2^{(k+1)})\leq T(P_1^{(k)},P_2^{(k)})$. The proposed model based on the updating rule (Algorithm 1) is decreasing monotonically. Since the optimization problem is convex, the objective function finally converges to the global optimal solution.

\section{Experiments}
 In this section, we conduct some experiments on three benchmark datasets, i.e. Pascal-Sentence \cite{cybernetics:Wei}, MIRFlickr \cite{ACM:Huiskes} and NUS-WIDE \cite{ACM:Chua}. Two widely-used evaluation protocols are introduced to evaluate our algorithm.\\

\begin{table}
\centering
\caption{Statistics of three standard datasets}
\label{Tab:1}       
\begin{tabular}{cccc}
\hline\noalign{\smallskip}
DataSets & Pascal-Sentence & MIRFlickr & NUS-WIDE \\
\noalign{\smallskip}\hline\noalign{\smallskip}
Data Set Size & 1000  & 16738 & 190420\\
Training Set Size & 600 & 5000 & 5212\\
Retrieval Set Size & 600  & 5000 & 5212\\
Query Set Size & 400 & 836 & 3475 \\
Num. of Labels & 20 & 24 &  21 \\
\noalign{\smallskip}\hline
\end{tabular}
\end{table}

\subsection{Datasets}
\textbf{Pascal-Sentence} consists of 1000 samples with image-text pairs. Each image is described in several sentences. The dataset is divided into 20 categories and each category includes 50 samples. We randomly select 30 samples from each category as the training set and the rest for the testing set. For each image, we employ a convolutional neural network to extract 4096-dimension CNN features. For text features, We utilize the LDA model to get the probability of each topic, and the 100-dimensional probability vector is used to represent text features.\\
\hspace*{0.35cm}\textbf{MIRFlickr} contains original 25,000 images crawled from the Flickr website. Each image and its associated textual tags is called an instance. Each instance is manually classified into some of the 24 classes. We only keep those instances whose textual tags appear at least 20 times and remove those instances without annotated labels or any textual tags. Each image is represented by a 150-dimensional edge histogram vector, and each text is represented as a 500-dimensional vector derived from PCA on the bag-of-words vector. We randomly select 5\% of the instances as the query set and 30\% of the instance as the training set.\\
\hspace*{0.35cm}\textbf{NUS-WIDE} is a subset sampled from a real-world web image dataset including 190,420 image-text pairs with 21 possible labels. For each pair, image is represented by 500-dimensional SIFT BoVW features and 1000-dimensional text annotations for text. The dataset contains 8,687 image-text pairs which are divided into two parts: 5,212 pairs for training and 3,475 pairs for testing.

\subsection{Experimental setting}
The proposed CKD in this paper is a supervised, kernel-based and correlation-based method. We compare our algorithm with the following methods:   CCA \cite{Neuralcomputation:Hardoon} ( Correlation-based method); KCCA \cite{ACM:Lisanti} (Kernel-based and Correlation-based method); ml-CCA \cite{ICCV:Ranjan} (Supervised and Correlation-based method); KDM \cite{Neurocomputing:Xu} (Supervised, Kernel-based and Correlation-based method). Besides, we compare our algorithm with the case where $\beta$ is set as zero. We tune the parameter $d$ in the range of \{10,20,30,40,50,60\}. $\lambda_1$ and $\lambda_2$ are two coefficients of the regularization term and they are fixed as 0.01 in experiments.  $\alpha_1$ and $\alpha_2$ are trade-off parameters in the objective function. We set the possible values of $\alpha_1$ and $\alpha_2$ in the range of \{1e-5,1e-4,1e-3,1e-2,1e-1,1,10,1e2,1e3,1e4,1e5 \} empirically. The best results are reported in this paper. Our experiments are implemented on MATLAB 2016b and Windows 10 (64-Bit) platform based on desktop machine with 12 GB memory and 4-core 3.6GHz CPU, and the model of the CPU is Intel(R) CORE(TM) i7-7700.
\subsection{Evaluation Protocol}
There are many evaluation metrics in the information retrieval area. We introduce two commonly used indicators, i.e. Cumulative Match Characteristic Curve (CMC) and Mean Average Precision (MAP). For a query $q$, the Average Precision (AP) is defined as E.q.(15)
\begin{equation}
AP(q) = \frac{1}{l_q}\sum_{m=1}^RP_q(m)\delta_q(m)
\end{equation}
where $\delta_q(m)=1$ if the result of position $m$ is right and  $\delta_q(m)=0$ otherwise; $l_q$ represents the correct statistics of top $R$ retrieval results; The average value of AP of all queries is called MAP which indicates a better performance with larger value.\\
\hspace*{0.35cm}CMC is the probability statistics that the true retrieval results appear in different-sized candidate lists. Specifically, if retrieval results contain one or more objects classified into the same class with query data, we think that this query can match true object. Assuming that the length of retrieval results is fixed as $m$, the rate of true match in all query is denoted as CMC$_{rank-m}$. 
\subsection{Results}
We perform some experiments for two typical retrieval tasks: Image query Text database and Text query Image database which are abbreviated as 'I2T' and 'T2I' respectively in this paper. I2T indicates known an image as a query, to retrieve the same semantic texts wit the image from the text database. By analogy, the meaning of T2I is as follows: Given a text, to search for images being similar content to the text from the image database. In this paper, the normalized correlation (NC) \cite{relatedwork9} which is an effective measurement strategy is adopted to measure the similarity among data. Table 1 shows the MAP results on Pascal-Sentence, MIRFlickr, and NUS-WIDE. As reported in Table 1, we can observe that CKD outperforms the compared methods. Specifically, CKD achieves average improvements of 6.14\%, 1.31\%, and 9.97\% over the best baselines on Pascal-Sentence, MIRFlickr, and NUS-WIDE respectively. Besides, we set the candidate list size in rank \{5,10,15,20,25,30\}. according to the CMC protocol, we conduct some experiments to further validate the effectiveness of CKD. Fig. 3 and Fig. 4 show the performance variation of all approaches with respect to different-sized candidate lists for I2T and T2I respectively. As illustrated in Fig. 3 and Fig. 4, our model achieves better performance than other approaches. The above experimental results on Pascal-Sentence, MIRFlickr, and NUS-WIDE indicate that CKD proposed in this paper is effective for cross-modal retrieval.
\begin{table}[h]
\centering
\caption{The MAP results on Pascal-Sentence, MIRFlickr and NUS-WIDE}
\label{Table 1}
\begin{tabular}{|c|c|c|c|c|}
\hline
Datasets&Approaches&I2T&T2I&Avg.\\
\hline
\hline
\multirow{6}{*}{Pascal-Sentence}&CCA&0.0501&0.0456&0.0479\\
\cline{2-5}
&KCCA&0.0376&0.0402&0.0389\\
\cline{2-5}
&ml-CCA&0.0422&0.0329&0.0376\\
\cline{2-5}
&CKD($\beta$=0)&0.0736&0.1300&0.1018\\
\cline{2-5}
&KDM&0.1729&0.1992&0.1861\\
\cline{2-5}
&CKD&\begin{bfseries}0.2143\end{bfseries}&\begin{bfseries}0.2806\end{bfseries}&\begin{bfseries}0.2475\end{bfseries}\\
\hline

\multirow{6}{*}{MIRFlickr}&CCA&0.5466&0.5477&0.5472\\
\cline{2-5}
&KCCA&0.5521&0.5529&0.5525\\
\cline{2-5}
&ml-CCA&0.5309&0.5302&0.5306\\
\cline{2-5}
&CKD($\beta$=0)&0.5602&0.5595&0.5599\\
\cline{2-5}
&KDM&0.5951&0.5823&0.5887\\
\cline{2-5}
&CKD&\begin{bfseries}0.6103\end{bfseries}&\begin{bfseries}0.5933\end{bfseries}&\begin{bfseries}0.6018\end{bfseries}\\
\hline

\multirow{6}{*}{NUS-WIDE}&CCA&0.3099&0.3103&0.3101\\
\cline{2-5}
&KCCA&0.3088&0.3174&0.3096\\
\cline{2-5}
&ml-CCA&0.2787&0.2801&0.2794\\
\cline{2-5}
&CKD($\beta$=0)&0.3170&0.3164&0.3167\\
\cline{2-5}
&KDM&0.3247&0.3118&0.3183\\
\cline{2-5}
&CKD&\begin{bfseries}0.4149\end{bfseries}&\begin{bfseries}0.4211\end{bfseries}&\begin{bfseries}0.4180\end{bfseries}\\
\hline
\end{tabular}
\end{table}

\begin{table}
\centering
\caption{The comparative MAP results on NUS-WIDE}
\label{Tab:3}       
\begin{tabular}{cccc}
\hline\noalign{\smallskip}
Methods & I2T  & T2I & Avg. \\
\noalign{\smallskip}\hline\noalign{\smallskip}
Multimodal DBN & 0.2013  & 0.2594 & 0.2303 \\
Bimodal-AE & 0.3271 & 0.3693 & 0.3482\\
Corr-AE & 0.3658 & 0.4172 & 0.3915\\
DCCA & 0.4844 & 0.5088 & 0.4966\\
CMDN & 0.4923 & 0.5151 & 0.5037 \\
ACMR & 0.5437 & 0.5376 & 0.5407 \\
Deep-SM & 0.6793 & 0.6930 & 0.6862\\
DCKD(Proposed) & 0.6827 & 0.7026 & 0.6927\\
\noalign{\smallskip}\hline
\end{tabular}
\end{table}

\begin{figure*}[htbp]
\subfigure[Pascal-Sentence]{
\includegraphics[width=.3\textwidth]{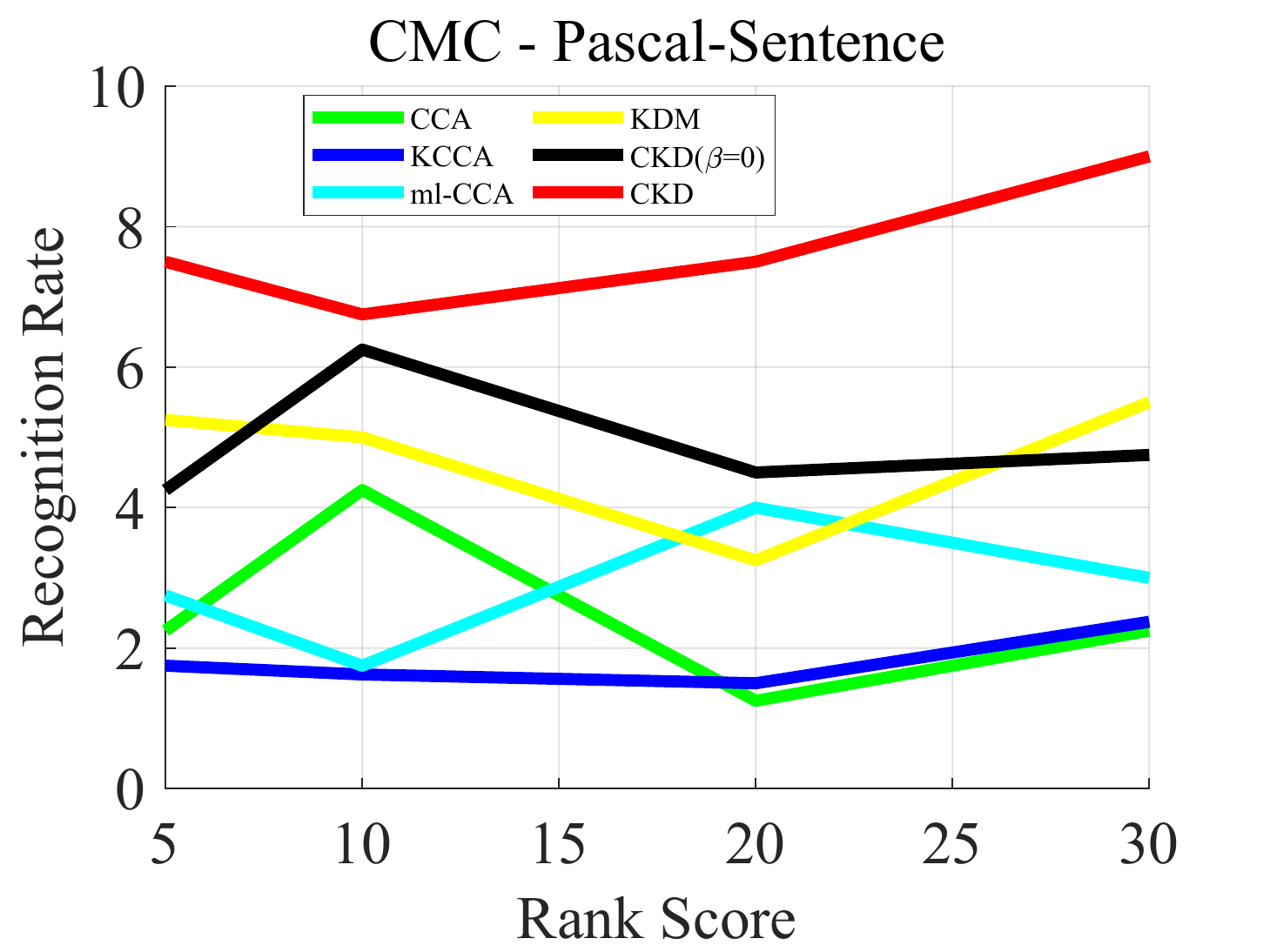}}
\subfigure[MIRFlickr]{
\includegraphics[width=.3\textwidth]{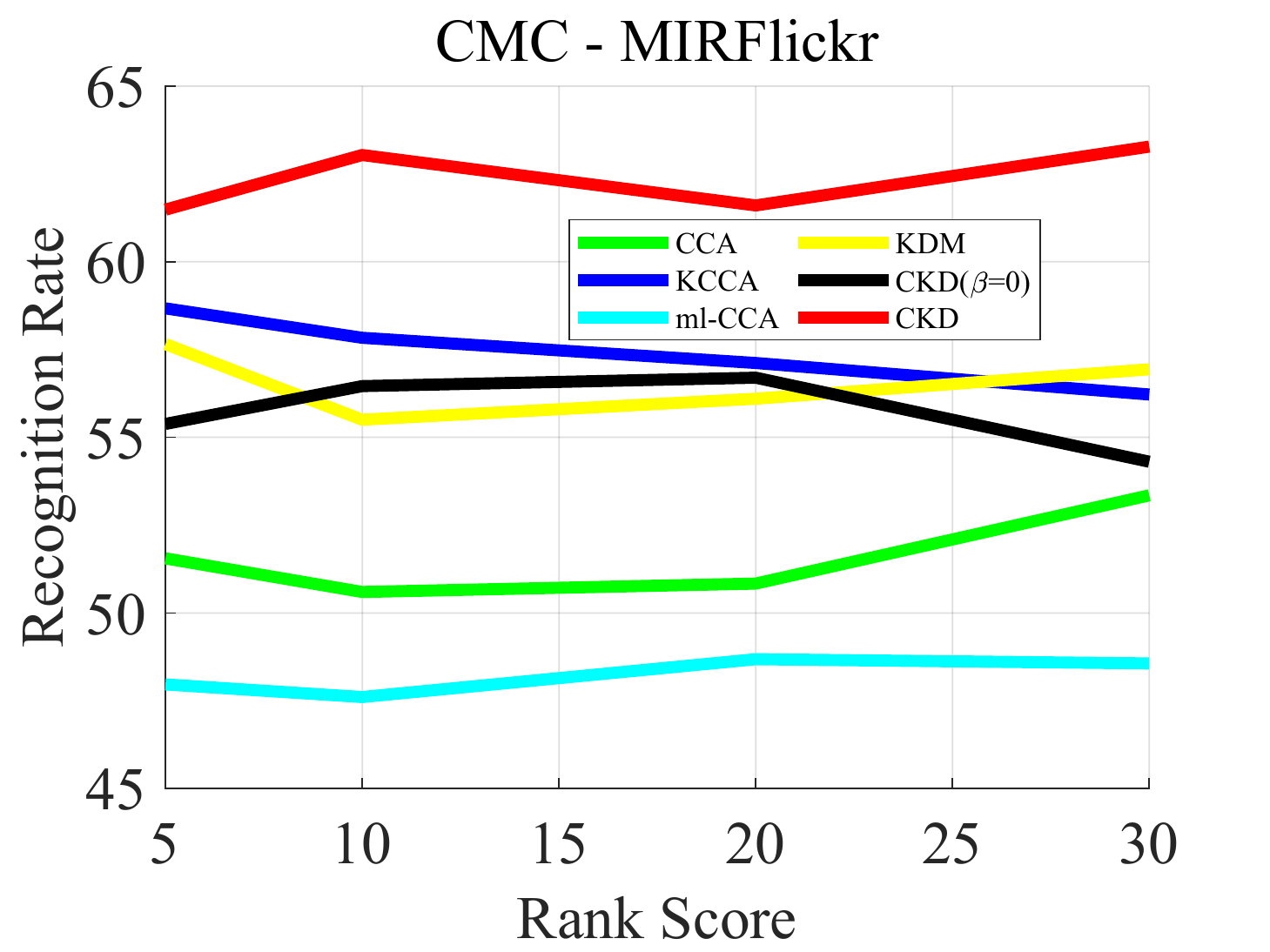}}
\subfigure[NUS-WIDE]{
\includegraphics[width=.3\textwidth]{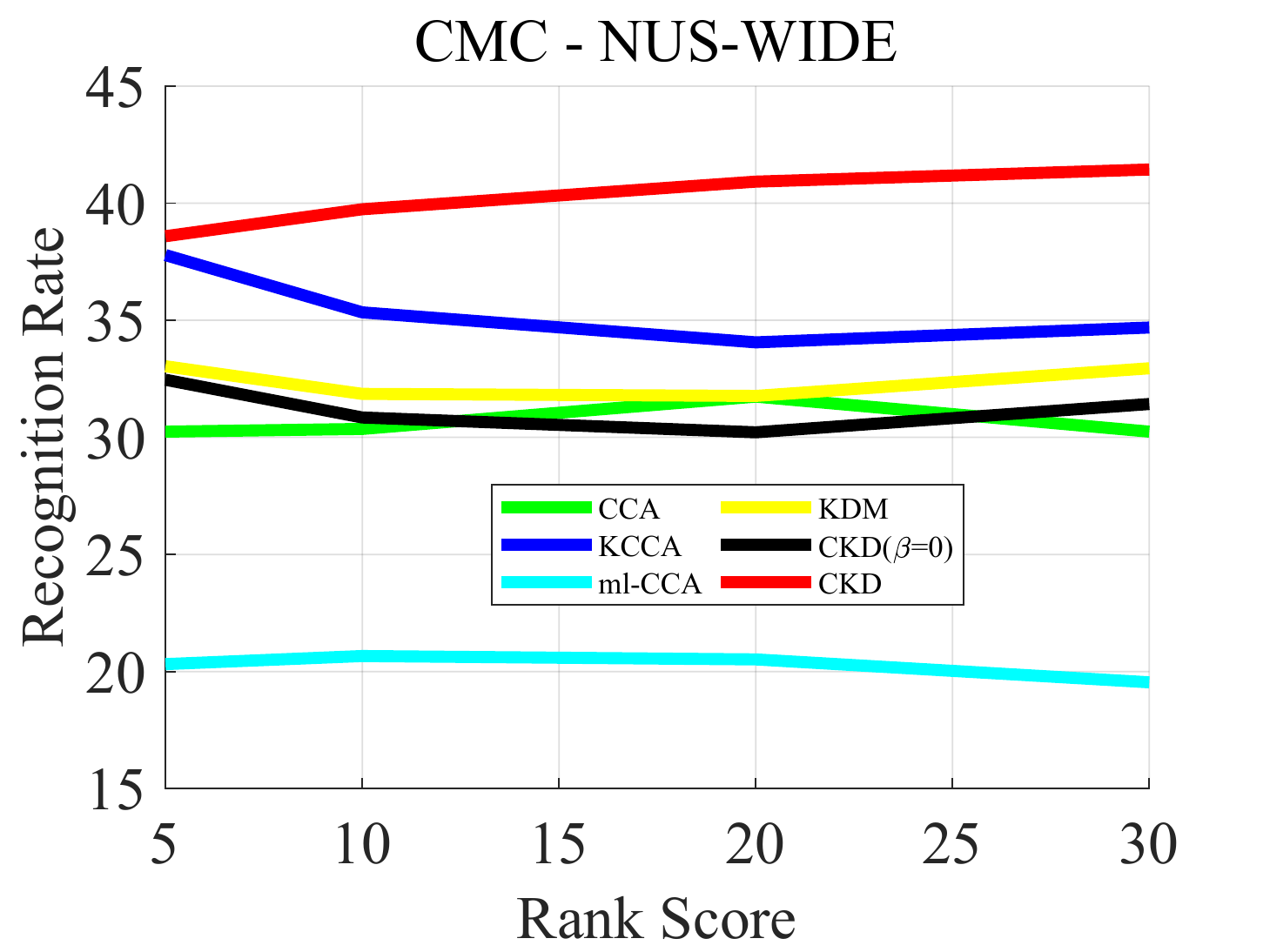}}
\caption{The CMC curve of all methods in terms of I2T on Pascal-Sentence(a), MIRFlickr (b), and NUS-WIDE (c).}
\label{Fig.3}
\end{figure*}

\begin{figure*}[htbp]
\subfigure[Pascal-Sentence]{
\includegraphics[width=.3\textwidth]{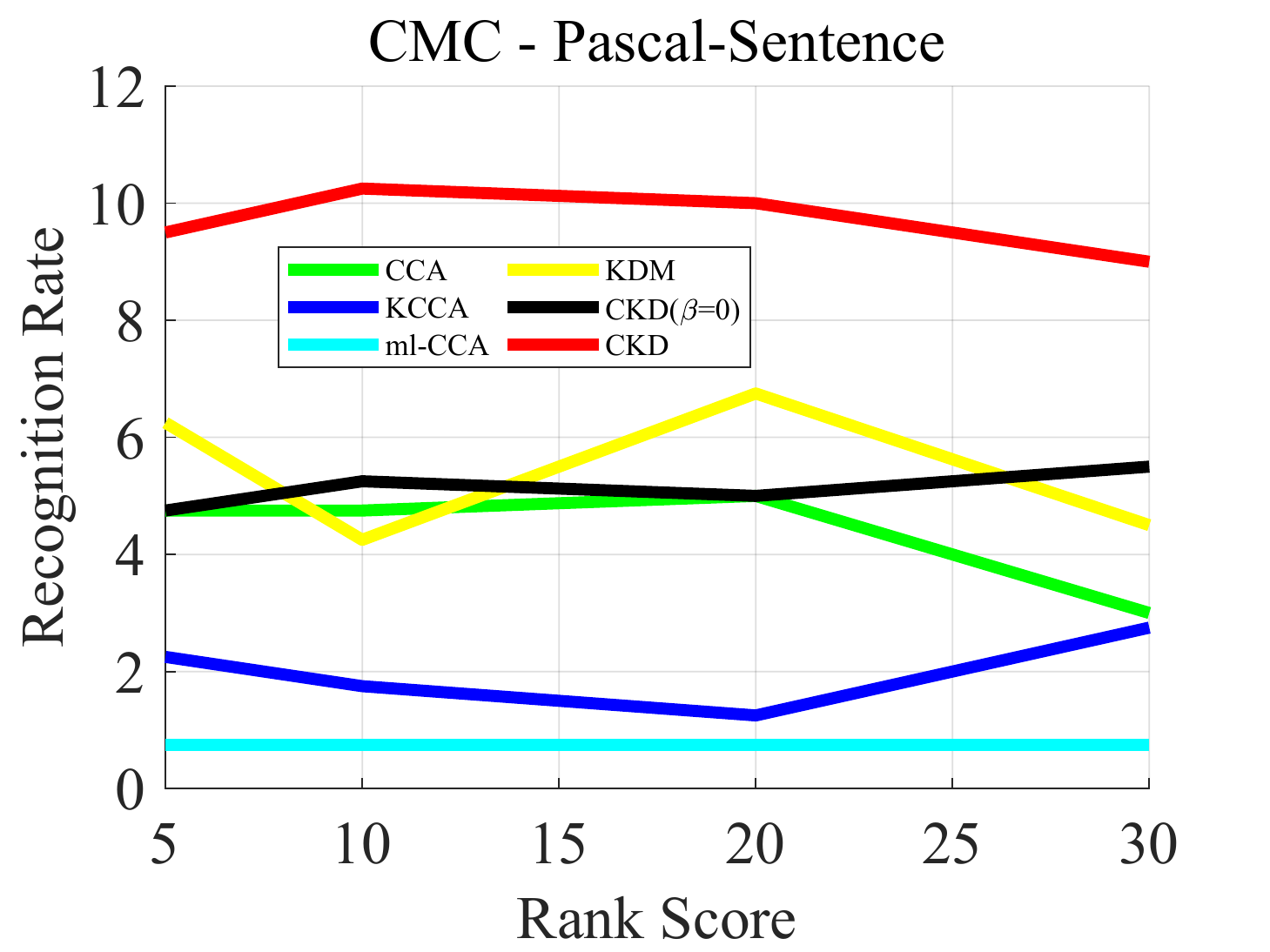}}
\subfigure[MIRFlickr]{
\includegraphics[width=.3\textwidth]{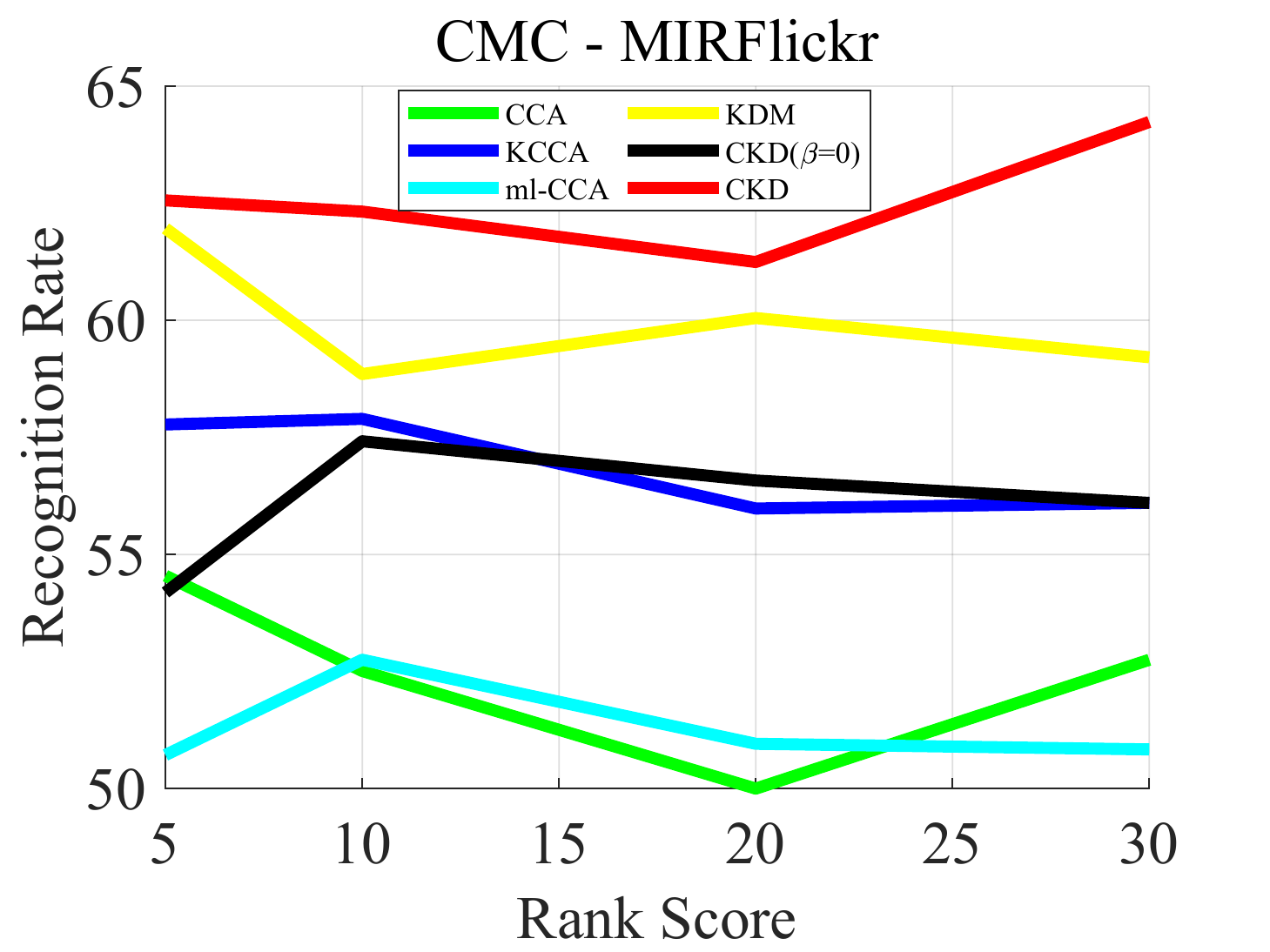}}
\subfigure[NUS-WIDE]{
\includegraphics[width=.3\textwidth]{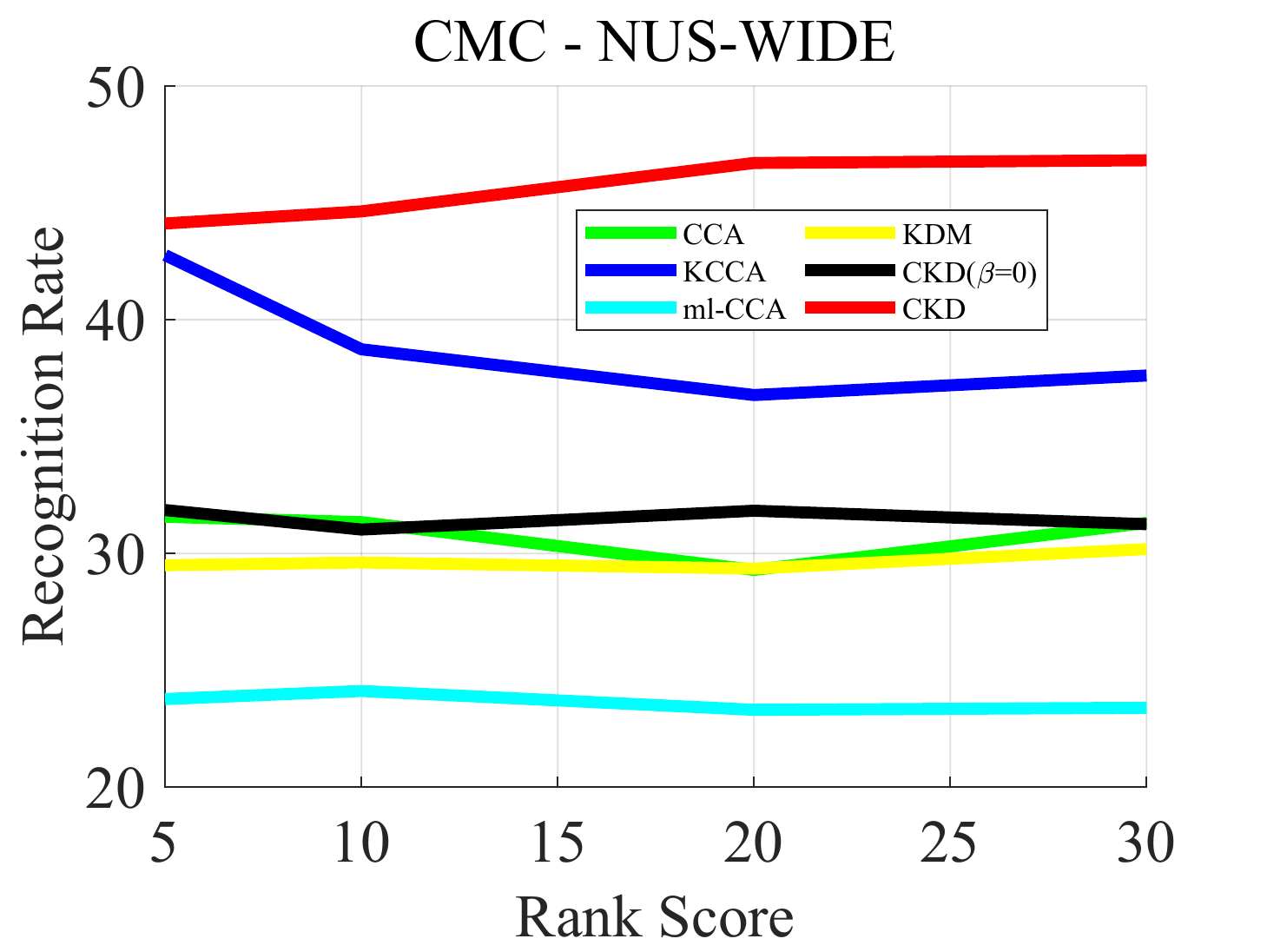}}
\caption{The CMC curve of all methods in terms of T2I on Pascal-Sentence(a), MIRFlickr (b), and NUS-WIDE (c)}
\label{Fig.4}
\end{figure*}

\begin{figure*}[htbp]
\subfigure[Pascal-Sentence (I2T)]{
\includegraphics[width=.3\textwidth]{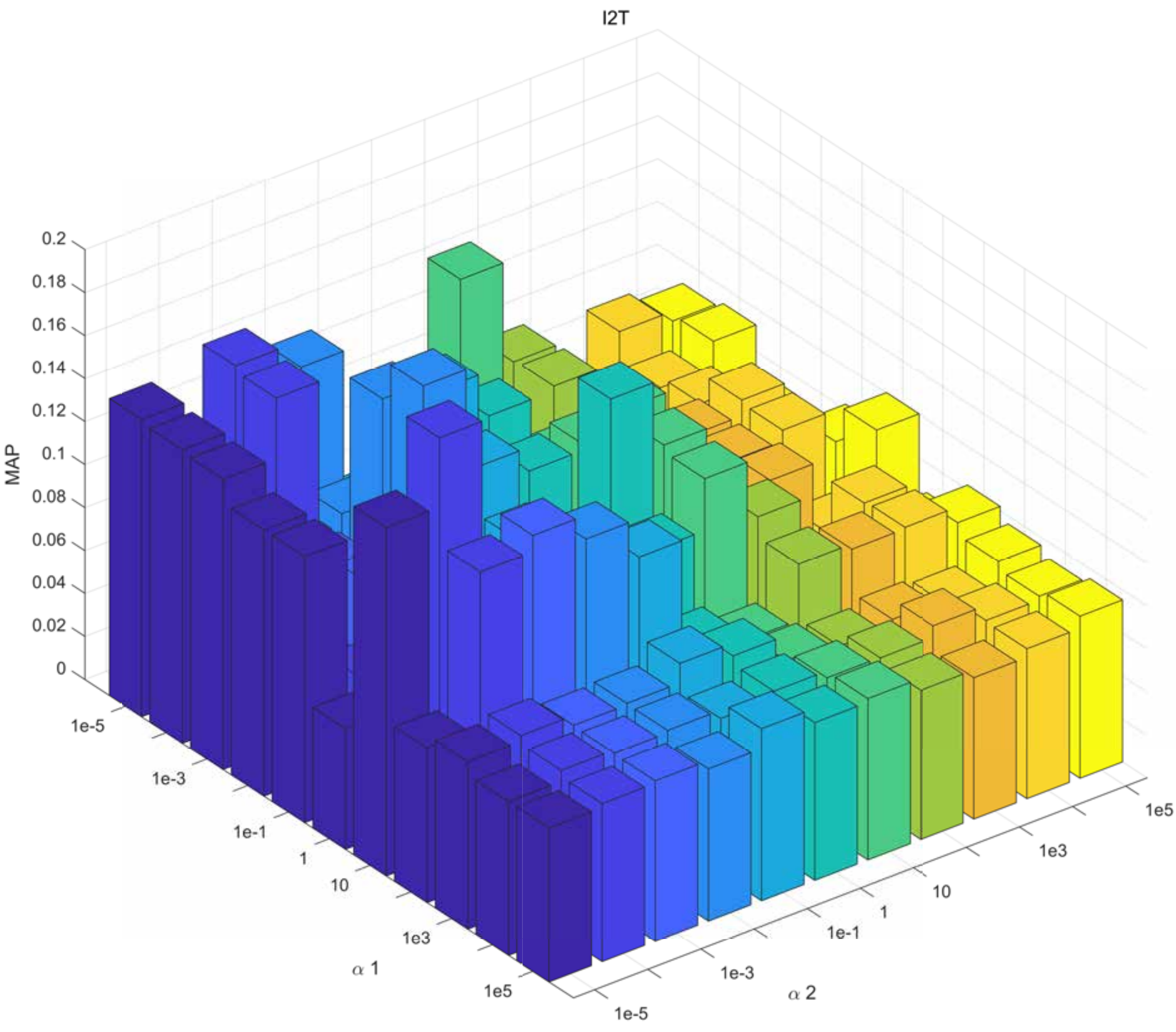}}
\subfigure[MIRFlickr (I2T)]{
\includegraphics[width=.3\textwidth]{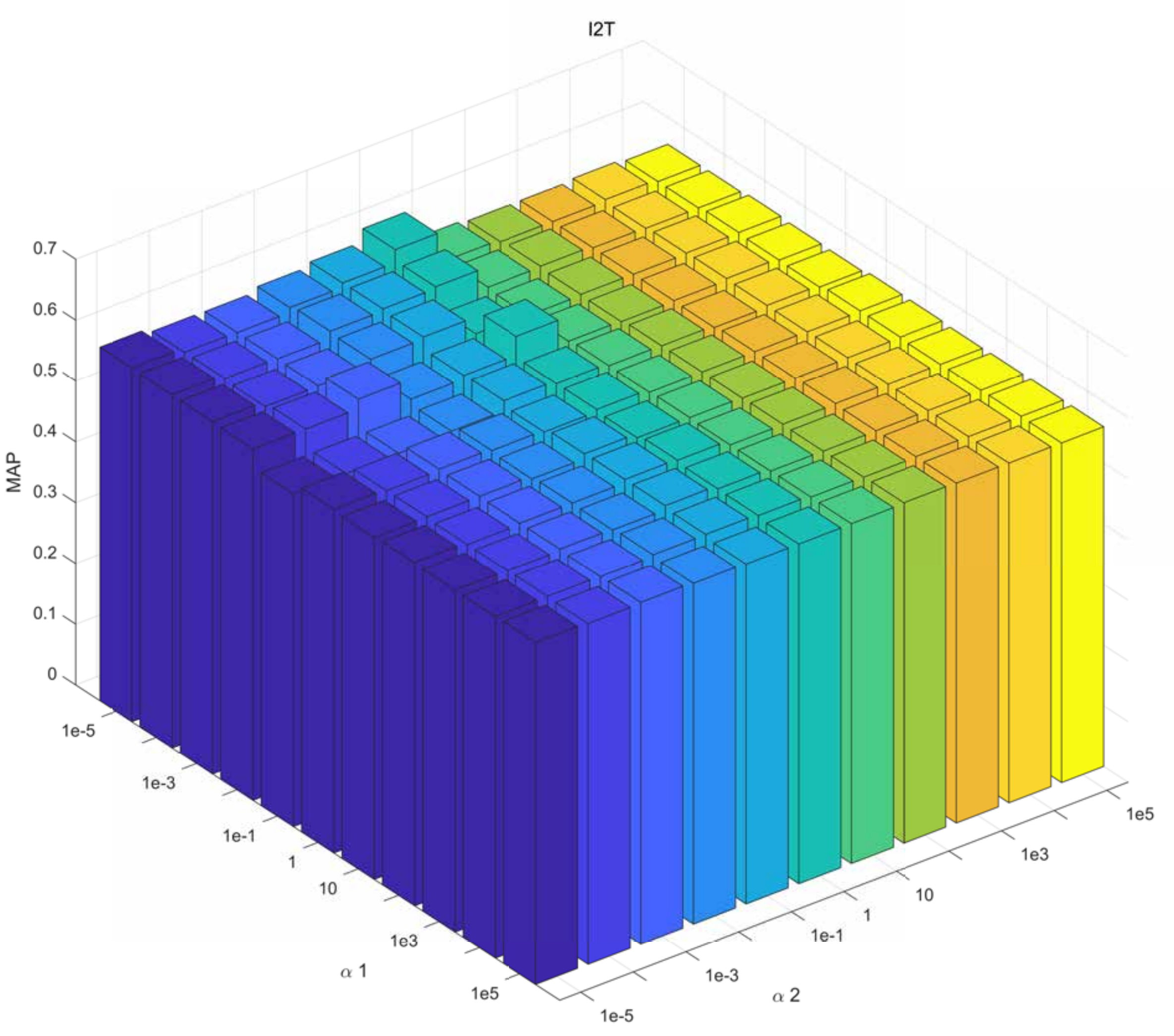}}
\subfigure[NUS-WIDE (I2T)]{
\includegraphics[width=.3\textwidth]{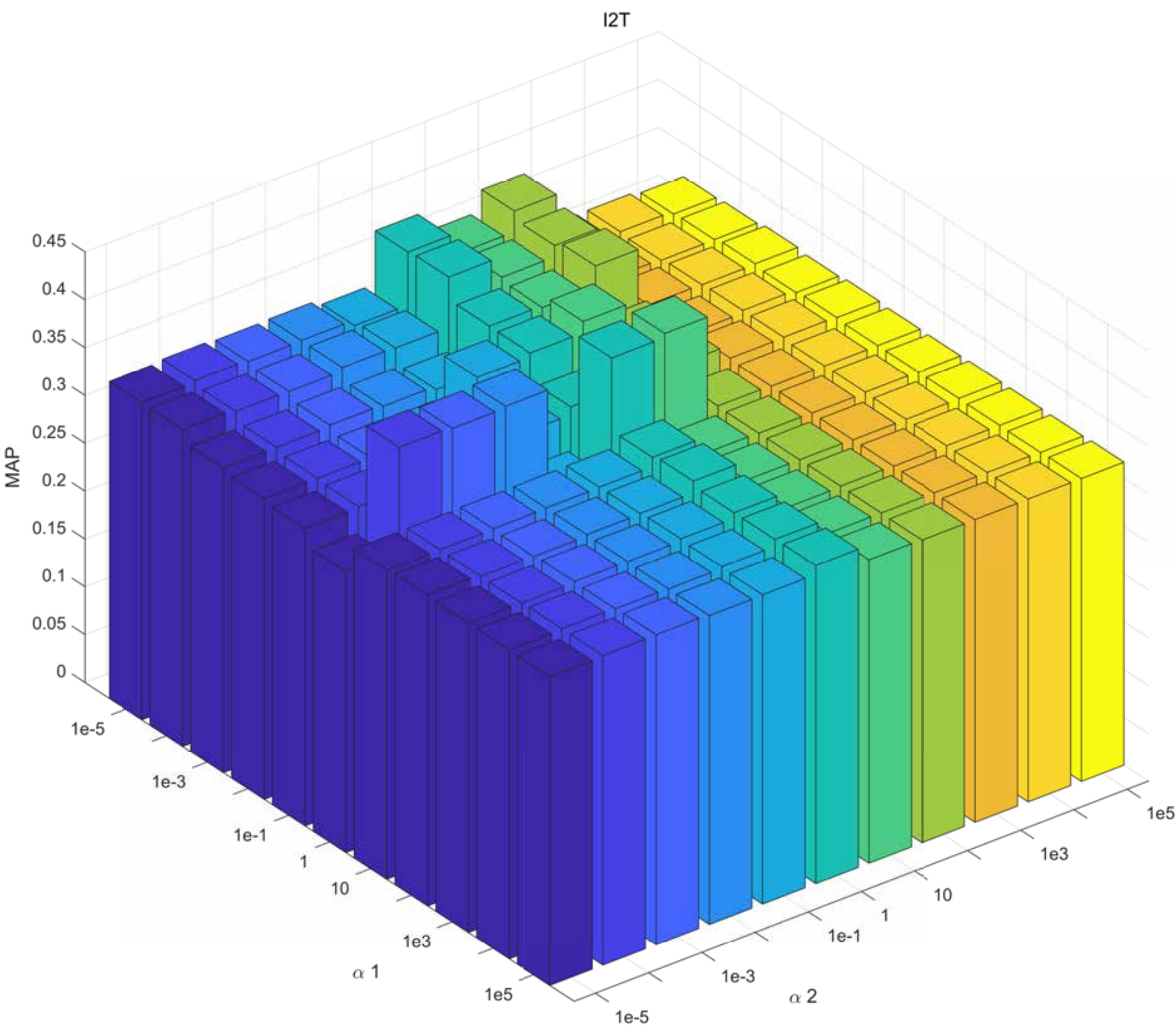}}
\subfigure[Pascal-Sentence (T2I)]{
\includegraphics[width=.3\textwidth]{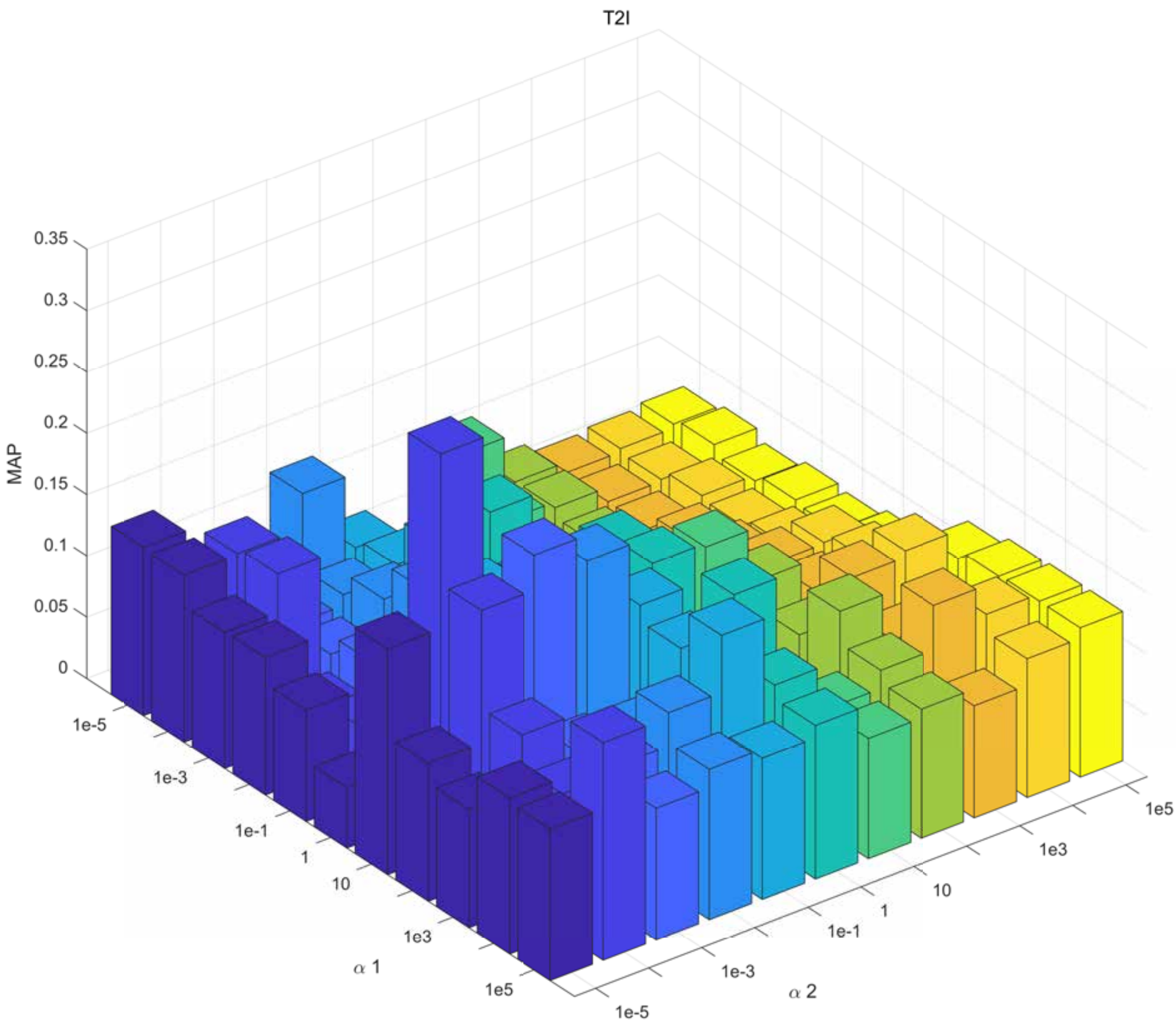}}
\subfigure[MIRFlickr (T2I)]{
\includegraphics[width=.35\textwidth]{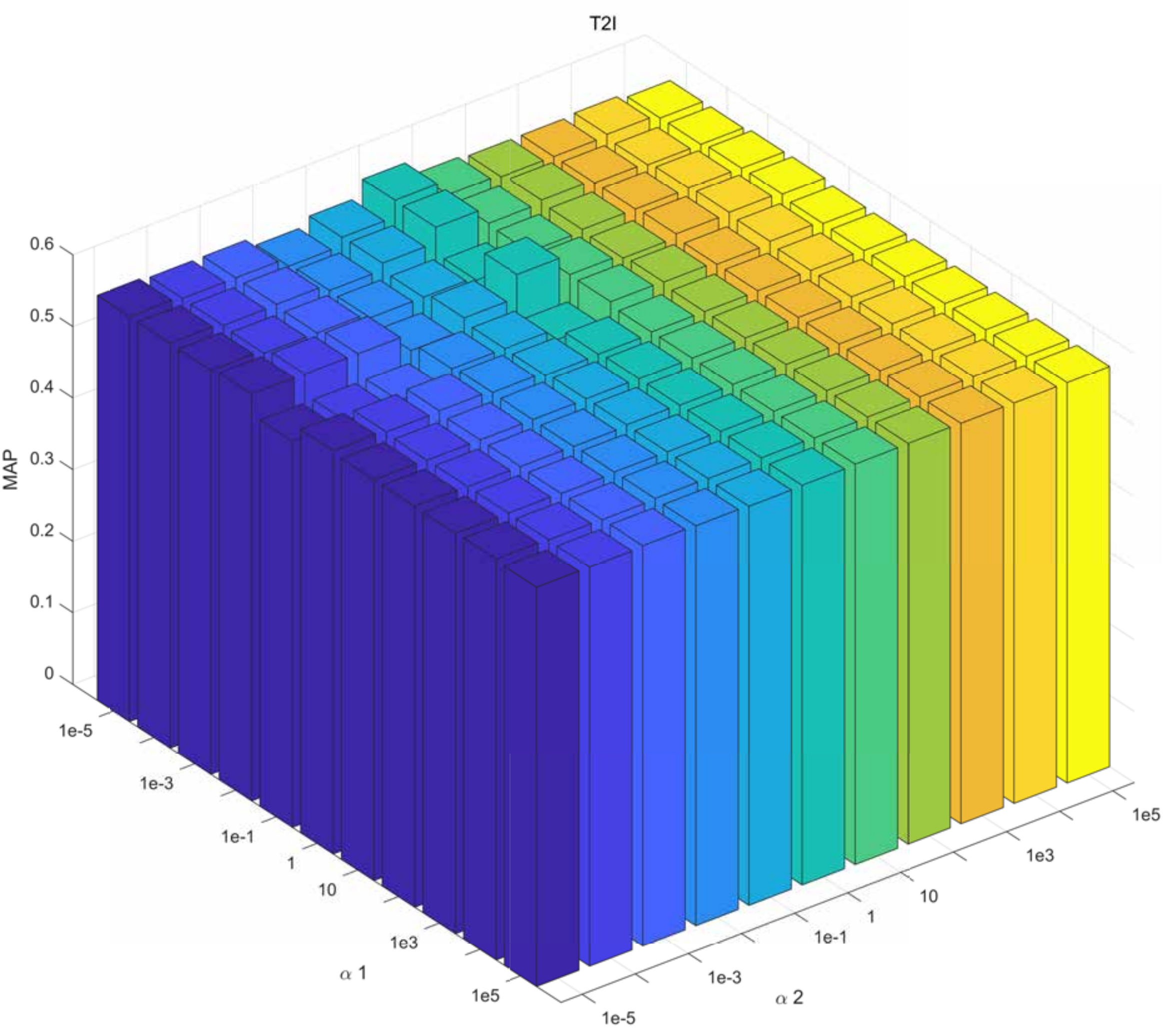}}
\subfigure[NUS-WIDE (T2I)]{
\includegraphics[width=.3\textwidth]{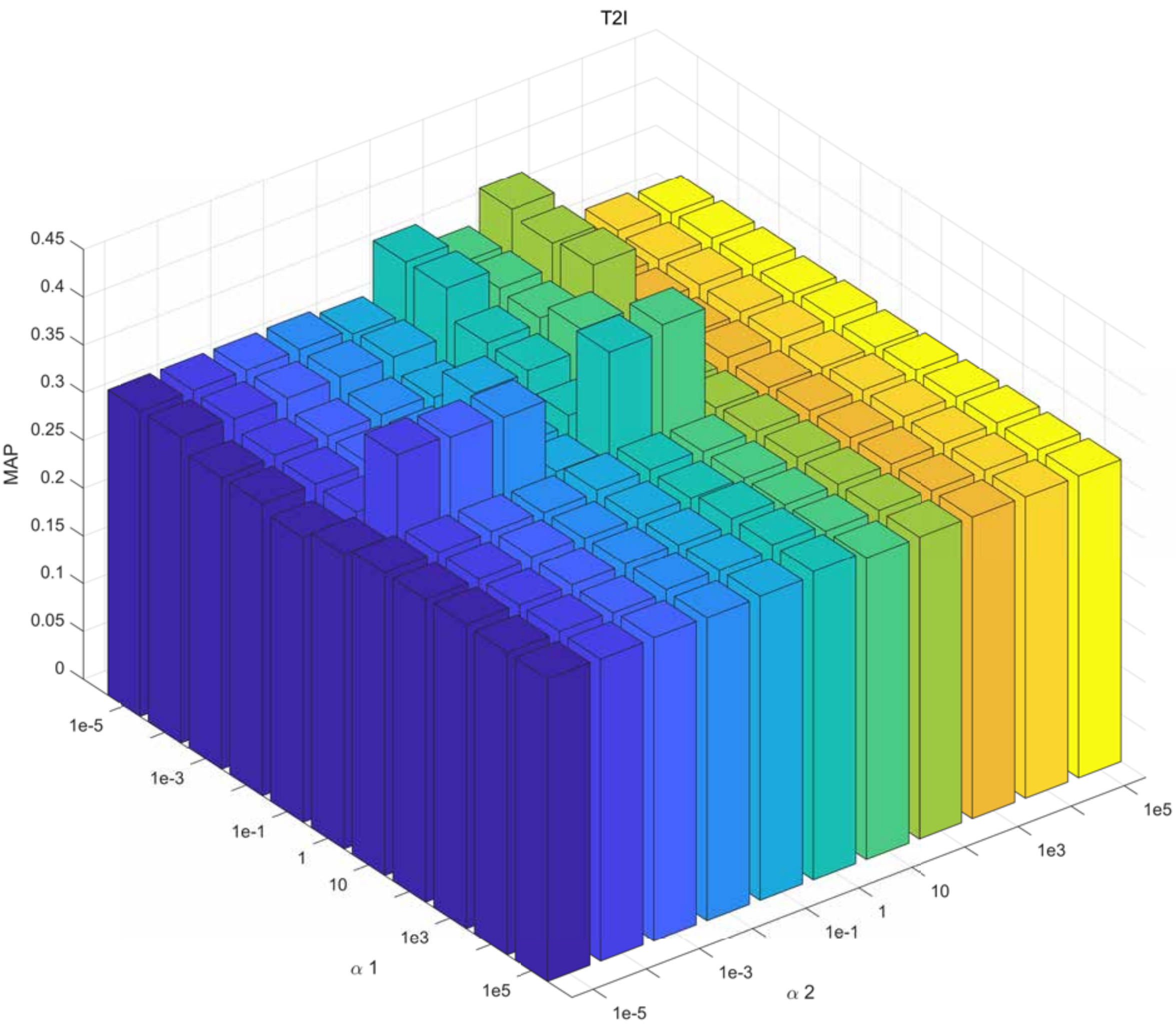}}
\caption{Performance variation of the CKD with respect to $\alpha_1$ and  $\alpha_2$ on all datasets.(a)  I2T on Pascal-Sentence.(b) I2T on MIRFlickr.(c) I2T on NUS-WIDE. (d)  T2I on Pascal-Sentence.(e) T2I on MIRFlickr.(f) T2I on NUS-WIDE.}
\label{Fig.5}
\end{figure*}

\begin{figure*}[htbp]
\subfigure[Pascal-Sentence]{
\includegraphics[width=.3\textwidth]{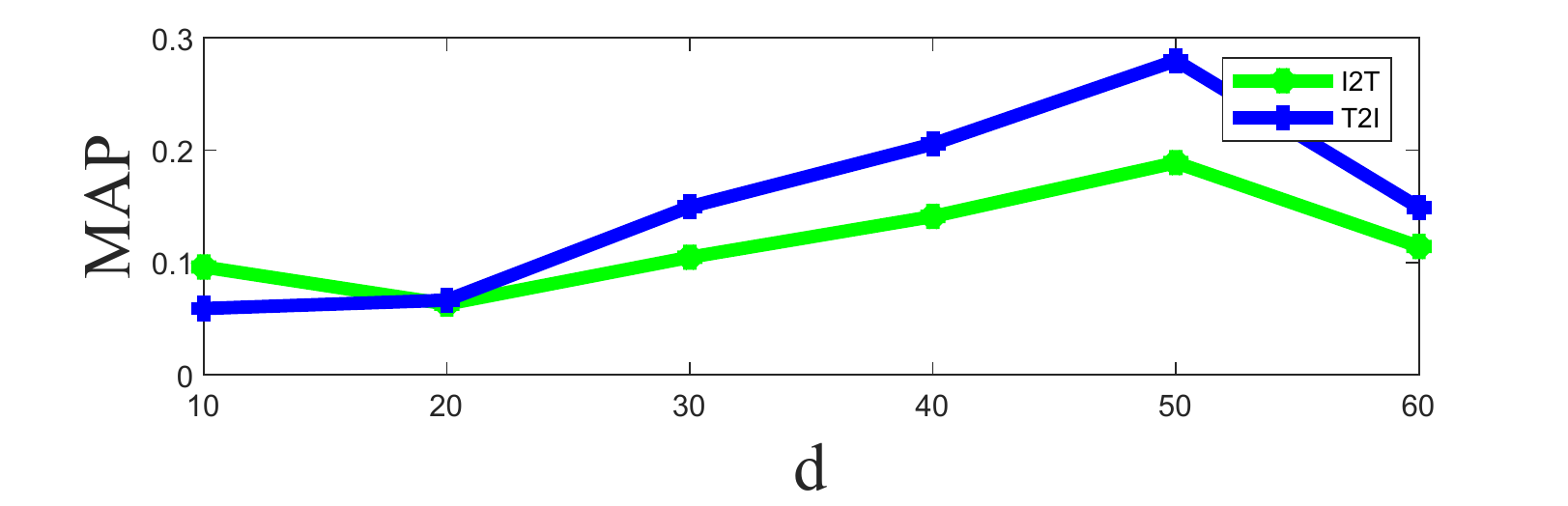}}
\subfigure[MIRFlickr]{
\includegraphics[width=.3\textwidth]{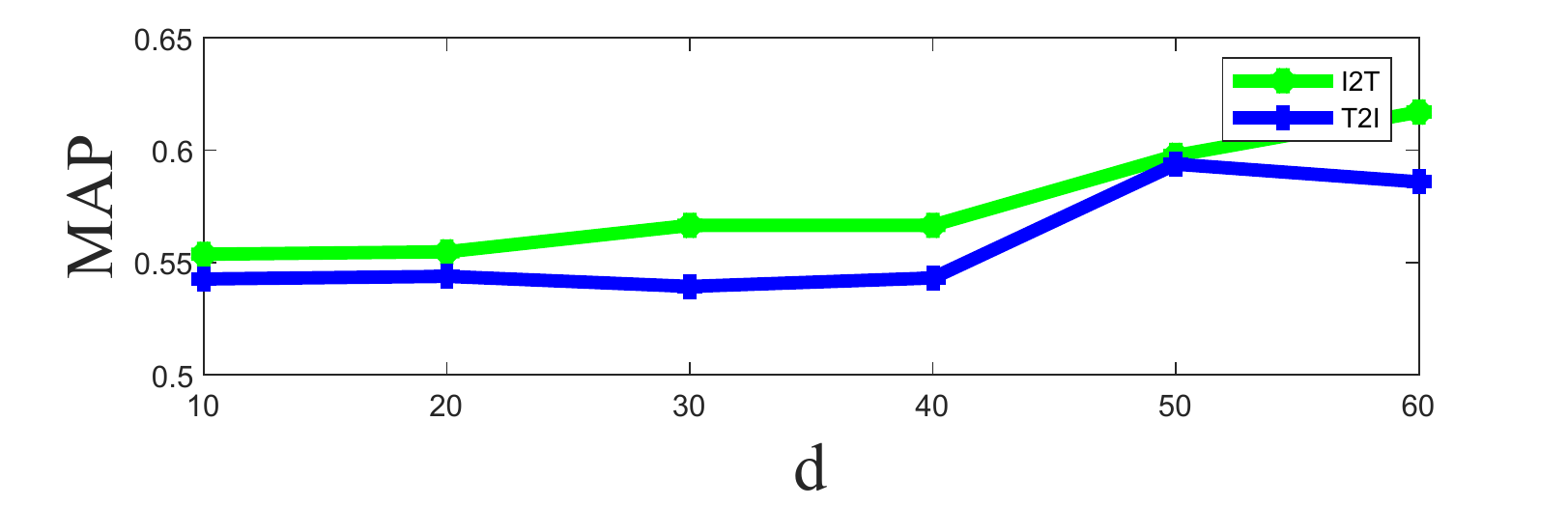}}
\subfigure[NUS-WIDE]{
\includegraphics[width=.3\textwidth]{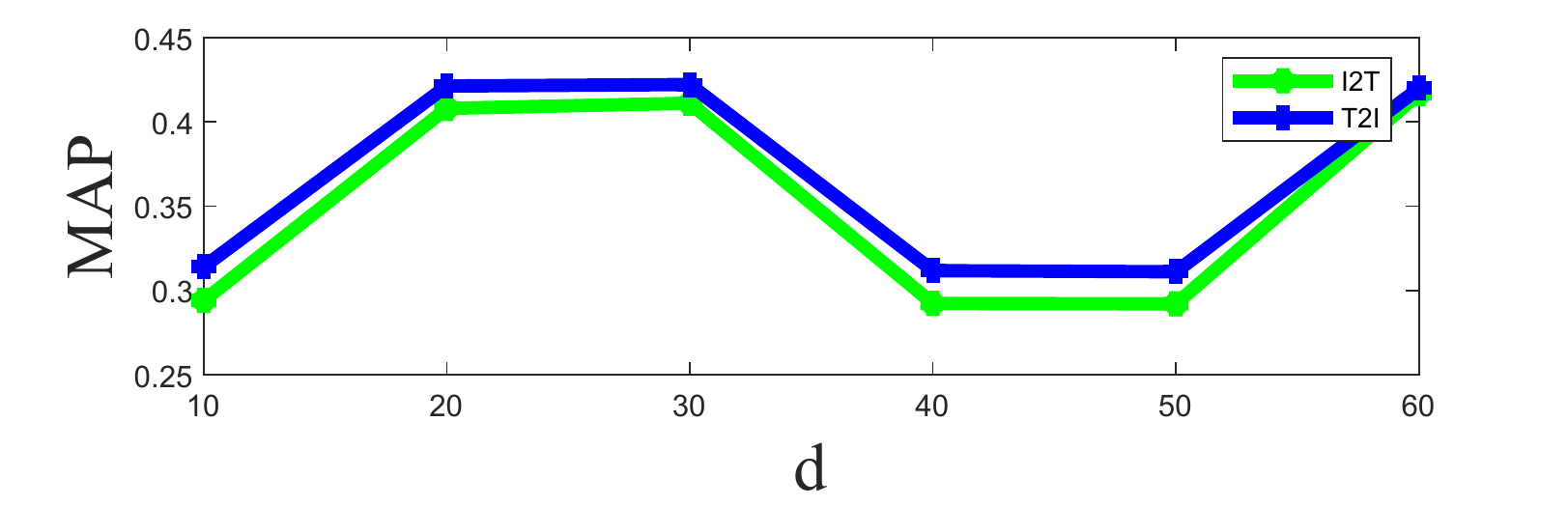}}
\caption{Performance variation of the CKD with respect to $d$ on Pascal-Sentence(a),MIRFlickr (b), and NUS-WIDE (c).}
\label{Fig.6}
\end{figure*}

\subsection{Discussion}
\subsubsection{Comparison with several classic subspace learning methods}
In our experiments, we perform some comparative experiments to validate the performance of the proposed CKD. On three standard benchmark datasets, our method outperforms several classic cross-modal subspace learning methods. CCA learns a latent common subspace by maximizing the pairwise correlations between two modalities. KCCA incorporates a kernel mapping based on CCA. ml-CCA, an extension of CCA, takes into account the semantic information in the form of multi-label annotations to learn shared subspace. Many CCA-like methods and CCA-based variants learn common subspace representation by maximizing the feature correlations between multi-modal data. Unlike these methods, KDM learns subspace representation by maximizing the kernel dependency. The performance of KDM is superior to CCA, KCCA, and ml-CCA which illustrates that the consistency between feature-similarity and semantic-similarity can help to learn more discriminative feature representation for cross-modal retrieval. Although KDM preserves the consistency of the similarity among samples for each modal, it does not ensure the discriminative feature representation in Hilbert space. To overcome this problem, the proposed CKD uses the label information to construct a semantic graph and hopes that the learned subspace representation can preserve the semantic structure. The framework of CKD includes two parts: kernel Correlation Maximization and Discriminative structure-preserving. To validate the contribution of each part to the final retrieval results, we conduct some ablation experiments on three benchmark datasets. KDM learns a subspace by maximizing the kernel correlation between multi-modal data, while CKD($\beta =0$) only considers preserving the semantic structure for each modal in the process of learning a common space. Obviously, CKD is superior to KDM and CKD($\beta =0$) in terms of retrieval precision, which manifests that the proposed CKD integrating the kernel correlation maximization and the discriminative structure-preserving can improve the retrieval performance.

\subsubsection{Performance Comparison of our model based on DNN features}
Deep neural networks have demonstrated their powerful ability to encode effective feature representation and have been successfully applied to the information retrieval field. In this section, we replace the original features of our model with the DNN features for the multi-modal input data, which is termed as DCKD. Specifically, the 4096 dimensional CNN visual features are extracted by the fc7 layer of VGGNet which is pre-trained on ImageNet. We introduce the word2vector neural network to generate a 300-dimensional vector to represent each text. The NUS-WIDE dataset has been widely used as a benchmark dataset on cross-modal retrieval by many researchers. We compare our DCKD method with 7 deep learning approaches on this dataset. The compared methods based on DNN include MultimodalDBN \cite{comparisondeep1}, Bimodal-AE \cite{comparisondeep2}, Corr-AE \cite{comparisondeep3}, DCCA \cite{relatedwork11}, CMDN \cite{comparisondeep4}, ACMR \cite{relatedwork12}, Deep-SM \cite{relatedwork13}. Table 3 shows the MAP results of our DCKD and the compared methods on NUS-WIDE. As shown in Table 3, we can see that our DCKD outperforms all compared methods. In Table 3, we can observe the following points: (1) Our method that leverages label information to model the intra-modal structure and the inter-modal correlation is very effective. (2) The performance of our model is improved significantly when using DNN features.

\subsection{Parameter sensitivity analysis}
In this section, we explore the impact of the parameters involved in the proposed model on retrieval precision.  As formulated in (7), the two parameters $\alpha_1$ and $\alpha_2$ control the weight of two modalities respectively. We observe the performance variation by tuning the value of $\alpha_1$ and $\alpha_2$ in the range of \{ 1e-5,1e-4,1e-3,1e-2,1e-1,1,10,1e2,1e3,1e4,1e5 \}. Fig. 5 plots the performance of CKD I2T and T2I as a function of $\alpha_1$ and $\alpha_2$. As shown in Fig.5, we can see that our model on Pascal-Sentence is more sensitive to  $\alpha_1$ and $\alpha_2$ than on MIRFlickr and NUS-WIDE. In addition, for the dimension $d$ of Hilbert space, we carry out experiments on Pascal-Sentence, MIRFlickr, and NUS-WIDE by changing the value of $d$ in the range of \{10,20,30,40,50,60\}. Fig.6 illustrates the MAP curve when $d$ changes in the candidate range, which reveals that CKD achieves the best performance when $d$ is set as 50,50 and 60 on Pascal-Sentence, MIRFlickr and NUS-WIDE respectively.

\begin{table}[h]
\centering
\caption{The comparison of training and testing time(seconds) on Pascal-Sentence dataset}
\label{Table 4}
\begin{tabular}{|c|c|c|c|c|c|c|}
\hline
Time $\backslash$ Methods&CCA&KCCA&ml-CCA&KDM&CKD\\
\hline
Training time&1.3462&21.2693&3410.0332&419.3401&410.7084\\
\hline
Testing time&0.0496&0.0903&0.2269&0.04699&0.04576\\
\hline
\end{tabular}
\end{table}

\subsection{Complexity analysis}
In this section, we discuss the complexity of the proposed CKD. The time complexity of Algorithm 1 is mainly on updating $P_1$ and $P_2$ by the eigenvalue decomposition on $Q_1$ and $Q_2$ respectively. In each iteration, the eigenvalue decomposition on $Q_v$ ($v=1,2$) costs $\mathcal{O}(d_v^3)$. If the algorithm converges after $t$ iterations, the total complexity of our model is $\mathcal{O}(td^3)$, where $d = \max(d_1, d_2)$. The cost of the training stage does not grow significantly with the increase in the size of the training set, which is flexible and adaptive enough for efficient large-scale retrieval tasks. Furthermore, we investigate the running time of the proposed CKD and other methods. Some experiments are conducted on Pascal-Sentence dataset to analyze the time consumption of the training and testing stage. The experimental results are summarized in Table 4. As shown in Table 4, we can observe that: (1) the running time of CKD is lower than ml-CCA whose complexity is $\mathcal{O}(n^2d^2 + d^3)$, where $n$ is the number of the training set. (2) In the testing stage, the proposed CKD is much more efficiently than the other approaches, which is very fast to search for information from the database.

\section{Conclusion}
In this paper, we present a novel method that integrates kernel correlation maximization and discriminative structure-preserving into a joint optimization framework. A shared semantic graph is constructed to make the subspace representation preserve the semantically structural information among data. Our model with the multiple supervision information facilitates to learn discriminative subspace representation for cross-modal retrieval. The experimental results on three publicly available datasets show that our approach is effective and outperforms several classic subspace learning algorithms.


%


\acknowledgments 
THE PAPER IS SUPPORTED BY THE NATIONAL NATURAL SCIENCE FOUNDATION OF CHINA(GRANT NO.61672265,U1836218), MURI/EPSRC/DSTL GRANT EP/R018456/1, AND THE 111 PROJECT OF MINISTRY OF EDUCATION OF CHINA (GRANT NO. B12018).

\end{spacing}
\end{document}